\documentclass[10pt,twocolumn,letterpaper]{article}

\usepackage[pagenumbers]{cvpr} %

\usepackage[dvipsnames]{xcolor}

\definecolor{cvprblue}{rgb}{0.21,0.49,0.74}
\usepackage[pagebackref,breaklinks,colorlinks,citecolor=cvprblue]{hyperref}

\usepackage{textpos}  %
\usepackage{url}  %
\usepackage{wrapfig}
\usepackage{makecell}
\newcolumntype{x}[1]{>{\centering\arraybackslash}p{#1pt}}
\newcolumntype{y}[1]{>{\raggedright\arraybackslash}p{#1pt}}

\newlength\savewidth\newcommand\shline{\noalign{\global\savewidth\arrayrulewidth
  \global\arrayrulewidth 1pt}\hline\noalign{\global\arrayrulewidth\savewidth}}
\newcommand{\tablestyle}[2]{\setlength{\tabcolsep}{#1}\renewcommand{\arraystretch}{#2}\centering\footnotesize}
\usepackage{pifont}
\newcommand{\cmark}{\text{\ding{51}}}
\newcommand{\xmark}{\text{\ding{55}}}
\newcommand{\mname}{S-Seg}
\usepackage{tabulary,multirow,overpic}
\usepackage{epigraph}
\usepackage{booktabs}
\usepackage{float}
\usepackage{nicefrac}       %
\usepackage{microtype}      %

\title{Exploring Simple Open-Vocabulary Semantic Segmentation}

\author{Zihang Lai\\
University of Oxford\\
{\tt\small zihang.lai@eng.ox.ac.uk}
}

\begin{document}

\maketitle
\begin{abstract}
Open-vocabulary semantic segmentation models aim to accurately assign a semantic label to each pixel in an image from a set of arbitrary open-vocabulary texts.
In order to learn such pixel-level alignment, current approaches typically rely on a combination of (i) image-level VL model (\eg CLIP), (ii) ground truth masks, and (iii) custom grouping encoders.
In this paper, we introduce \mname{}, a novel model that can achieve surprisingly strong performance \textbf{without} depending on any of the above elements.
\mname{} leverages pseudo-mask and language to train a MaskFormer, and can be easily trained from publicly available image-text datasets. 
Contrary to prior works, our model directly trains for pixel-level features and language alignment.
Once trained, \mname{} generalizes well to multiple testing datasets without requiring fine-tuning. 
In addition, \mname{} has the extra benefits of scalability with data and consistently improvement when augmented with self-training. 
We believe that our simple yet effective approach will serve as a solid baseline for future research. 
Our code will be released at \url{https://github.com/zlai0/S-Seg}.
\end{abstract}

\section{Introduction}

Open-vocabulary semantic segmentation presents a unique challenge as it requires assigning accurate semantic labels to each pixel in an image using arbitrary open-vocabulary texts, rather than a fixed set of classes. This means that the model must be able to segment and classify any arbitrary categories expressed in language. Achieving this requires  a robust, pixel-level alignment between images and textual descriptions, which enables accurate association of each pixel with the most relevant class from a dynamically provided set of textual categories. 

\begin{figure}[t]
    \centering
        \caption{\textbf{\mname{} result on a web image.} Our goal is to segment everything, including fictional characters like \emph{minions}.}

    \includegraphics[width=\linewidth]{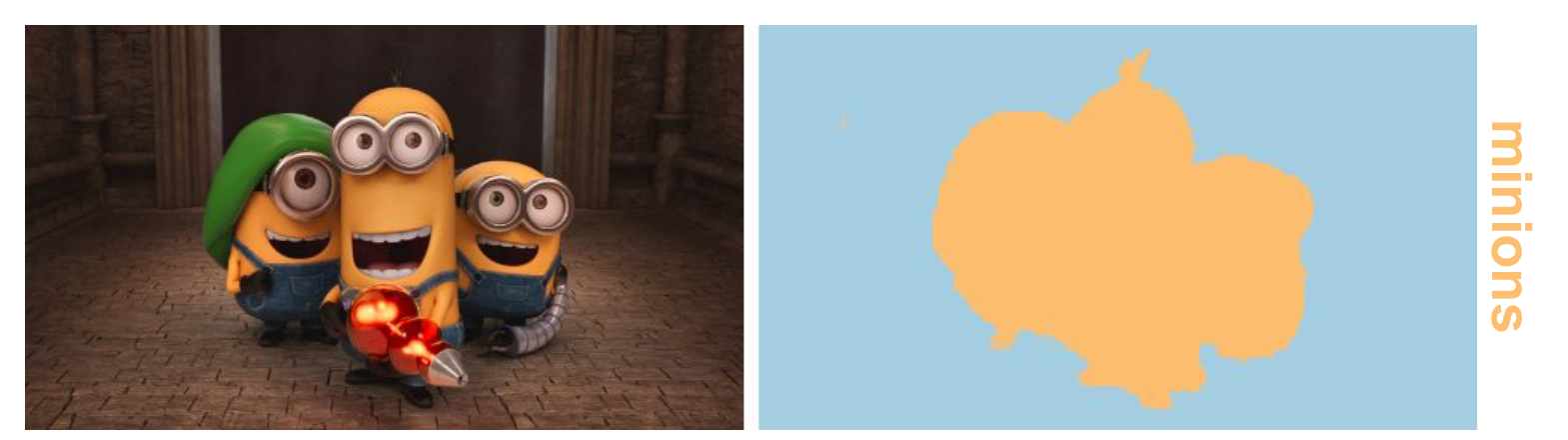}
    
    \label{fig:teaser}
\end{figure}

\begin{figure}[t]
    \centering
    \includegraphics[width=\linewidth]{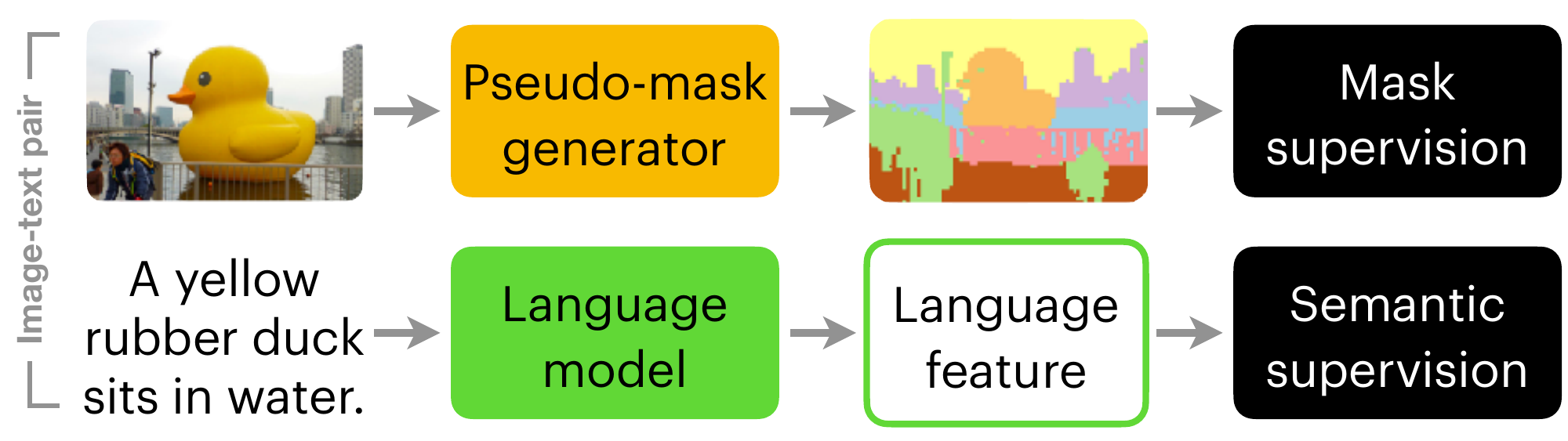}

    \caption{\small Our \textbf{\mname} framework leverages pseudo-mask and language to train a MaskFormer. We show that our method of directly training for pixel-level feature and language alignment yields superior results.}
    \vspace{-0.1in}
    \label{fig:illustration}
\end{figure}

\begin{figure*}[t]
    \centering
        \vspace{-1.5em}
    \includegraphics[width=0.95\linewidth]{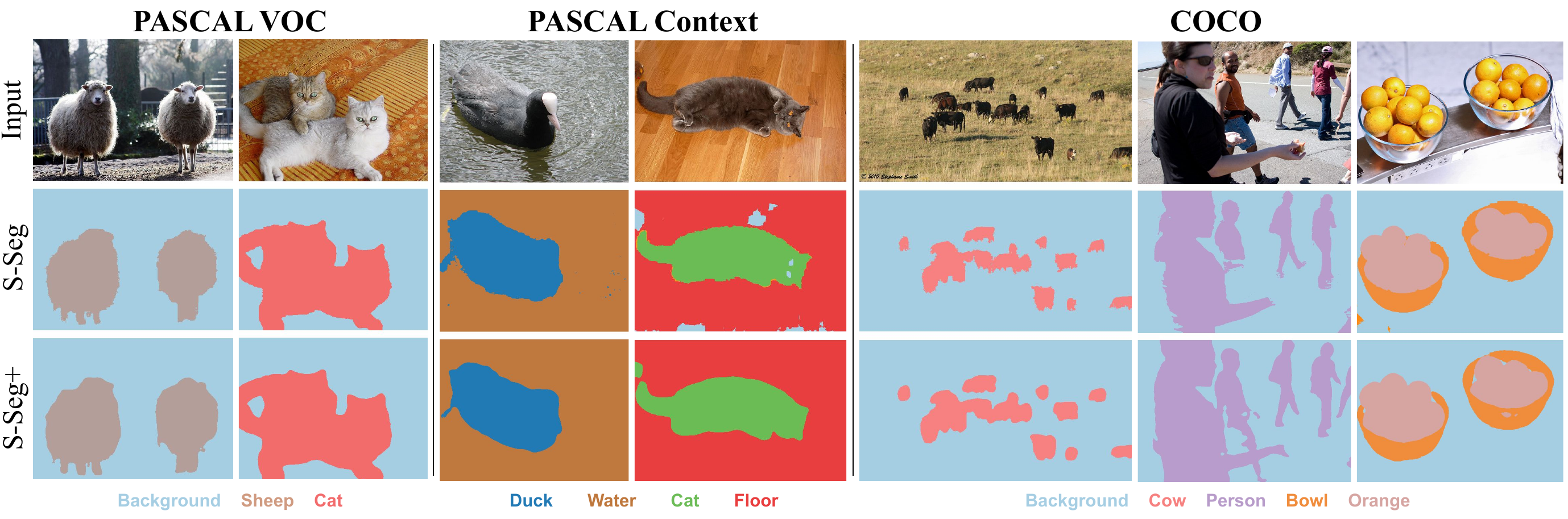}
    \caption{\small \textbf{Qualitative results of \mname{}, evaluated using \emph{all} dataset classes as queries.} 
    Our model copes with challenging situation, such as overlapping objects (col. 2) and small objects (col. 5). Our model is also capable of handling ``stuff" categories such as water and floor (col. 3, 4). Moreover, our \mname{}+ model is able to correct small errors observed in the \mname{} method (col. 4). Finally, in the COCO dataset, which featured a significantly higher number of objects, our model is still able to achieve high accuracy in its predictions.}
    \vspace{-1em}
    \label{fig:qualitative}
\end{figure*}

A primary obstacle in this domain is that it is impossible to construct datasets that provide pixel-level annotations for \emph{all} possible labels. This limitation often results in the adoption of weakly-supervised or semi-supervised learning approaches. Current methods typically rely on a combination of strategies to learn the required pixel-level alignment. One common tactic is adapting existing Vision-Language (VL) models, which are initially trained for image-level alignment (\eg, CLIP~\cite{radford2021learning}), to perform at the pixel level. 
Another strategy involves training models on ground truth masks that are annotated for a select number of \emph{seen} classes, thereby encouraging the model to extrapolate its learning to novel \emph{unseen} classes. Furthermore, specialized models such as GroupViT~\cite{xu2022groupvit} and OVSegmentor~\cite{xu2023learning}, which are explicitly designed for open-vocabulary segmentation, are being explored. These models typically group similar pixels within the image encoder based on their features, employing a hierarchical approach, enhancing the model's ability to understand the image at multiple granularities. 

In this paper, we report a model that can work surprisingly well with \emph{none} of the above strategies. Our approach, named \mname{}, is built on
top of a standard MaskFormer model. Our model directly trains for pixel-level feature and language alignment, using \emph{neither} existing large image-level alignment models like CLIP~\citep{radford2021learning} \emph{nor} manually annotated segmentation or classification labels.

One of the biggest challenges we face is finding the right supervision since annotated masks and labels are not available. To address this issue, we propose to leverage \emph{pseudo-masks} and \emph{language} to supervise MaskFormer. Our strategy involves using a pseudo-mask generator to provide class-agnostic mask supervision by generating pseudo ground truth masks. We adopt a simple design that clusters image representations obtained through self-supervised representation learning methods like DINO~\citep{caron2021emerging}. Our experiments demonstrate that this approach delivers exceptional performance, which is essential for high-quality supervision, as well as rapid processing speed, which is necessary for efficient training. In addition, we use noisy web texts to provide semantic supervision. The image-text dataset contains a wide range of concepts and has demonstrated impressive zero-shot classification results~\citep{radford2021learning}. We utilize a straightforward image-text contrastive loss, which has proven to be highly effective. Once trained, our model generalizes well to new categories without requiring fine-tuning.

 \mname{} is a simple and effective model that can be trained using publicly available image-text datasets, such as Conceptual Captions~\citep{sharma2018conceptual, changpinyo2021conceptual}. This makes it easy to reproduce and extend for further research. 
The \mname{} framework is also designed to be flexible with easily replaceable submodules. We prioritize \emph{simplicity} in our subcomponent selection to focus on the general design of our framework, while remaining open to more advanced techniques that could result in further improvements.

We conducted a thorough evaluation of \mname{} using multiple benchmark datasets, and we show that our method achieve competitive results on three widely tested benchmarks (Pascal VOC, Pascal Context, and COCO). In addition, pseudo-mask and language provide scalable supervision and our model consistently improves in performance as more data became available. Finally, we find adding an additional self-training step leads to an even greater improvement to our model, with an average increase of 5.5\% mIoU over three datasets, highlighting the potential for further improvement of our approach.

Our simple solution suggests that the reliance on complex models and extensive ground truth data in open-vocabulary semantic segmentation may be reduced, leading to a more streamlined and accessible framework for future developments in the field, and we hope our exploration can serve as a solid baseline for future research.

\begin{figure}[t]
    \centering
    \includegraphics[width=0.95\linewidth]{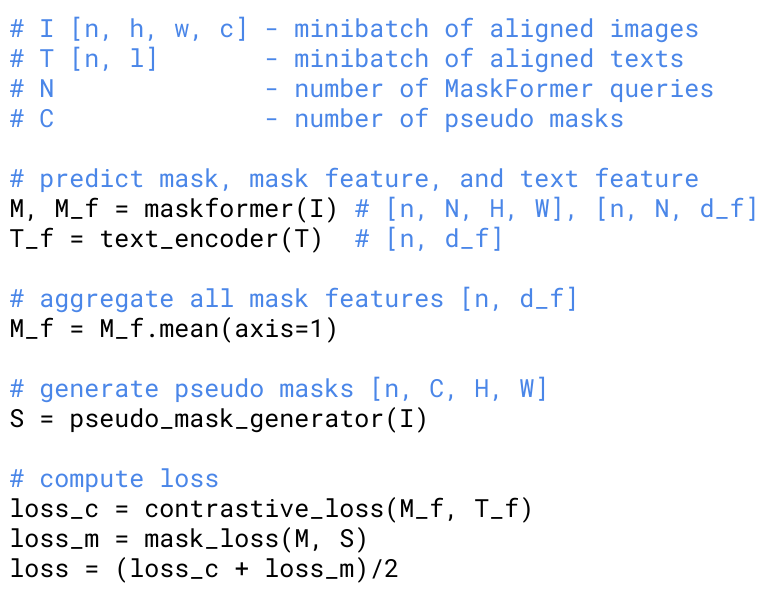}
    \caption{\textbf{Pseudocode for training \mname{} with image-text pairs.}}
    \vspace{-1em}
    \label{fig:pseudocode}
\end{figure}

\section{Related work}
\begin{figure*}[t]
    \centering
    \vspace{-1.5em}
    \includegraphics[width=0.9\linewidth]{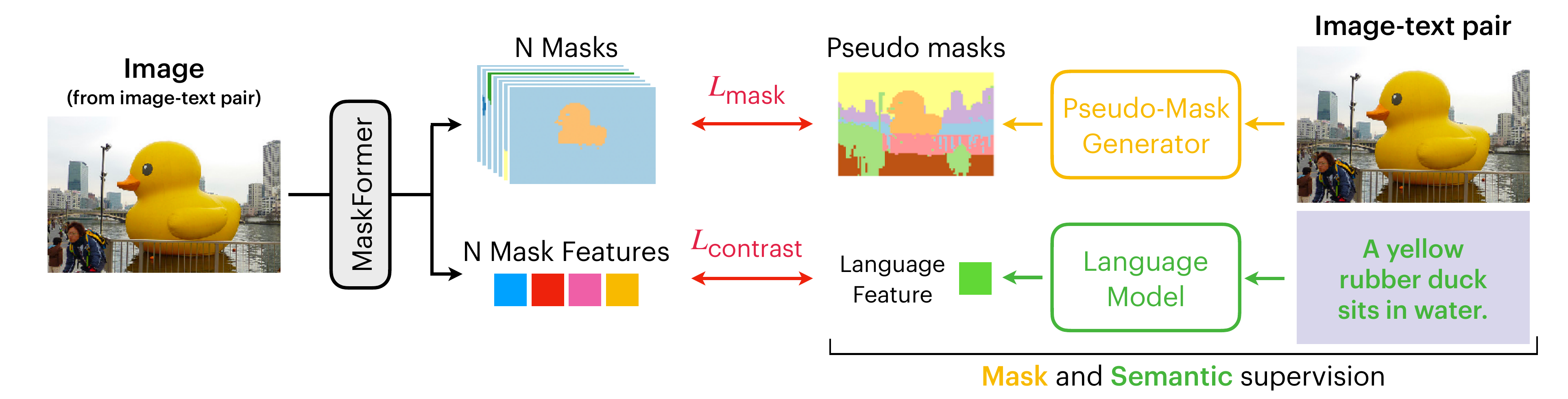}
    \vspace{-0.8em}
    \caption{\small \textbf{Overview of \mname{}.} A MaskFormer model computes masks and mask features from an image input. A pseudo-mask generator produces segmentation maps to supervise mask predictions, while a text that describes the image, encoded by a language model trained together with the MaskFormer, provides supervision for mask features using image-text contrastive loss. }
    \vspace{-1em}
    \label{fig:main}
\end{figure*}

\textbf{Open-vocabulary segmentation.}
The earliest efforts to employ language for image segmentation can be traced back to Duygulu et al.'s seminal work~\citep{duygulu2002object}, where the authors tackled image segmentation by framing it as a machine translation problem. 
Current approaches leverage a combination of strategy to learn pixel-level image-text alignment. 

\emph{Adapting image-level vision-language models.} The first strategy involves the adapting pretrained vision-language models, originally designed for image-level alignment, to the more granular task of pixel-level alignment. This strategy is widely adopted in open-vocabulary methods~\cite{zhou2022extract, ranasinghe2022perceptual,li2022language,ghiasi2022scaling,han2023global,han2023open,liang2023open,xu2023side,luo2023segclip,xu2023learning,cha2023learning, xu2023open, wu2023diffumask, karazija2023diffusion}. These works vary in their methods of refining image-level models for finer alignment tasks.

MaskCLIP~\cite{zhou2022extract} demonstrates modifying the CLIP image encoder can significantly enhance its pixel-level alignment capabilities without requiring retraining. TCL~\cite{cha2023learning} employs CLIP for initial text-to-image region grounding, followed by contrastive learning to refine the alignment between the text embedding and the grounded region. OpenSeg~\cite{ghiasi2022scaling} fine-tunes ALIGN~\cite{pmlr-v139-jia21b}  using a grounding loss~\cite{gupta2020contrastive} to better align words in captions to segmentation masks. OpenSeg and DiffuMask~\cite{wu2023diffumask} also explored the use of pseudo-masks. The primary distinction lies in their dependency on different sources for learning; OpenSeg uses annotated segments while DiffuMask employs masks generated through diffusion. In contrast, our method is entirely learned from pseudo masks. 
Also, our mask generator is entirely self-supervised, whereas their mask generator is fully-supervised.

\emph{Ground truth masks.} Another effective strategy~\cite{li2022language,ghiasi2022scaling,ding2022decoupling,han2023global,han2023open,liang2023open,xu2023side} involves training models using ground truth masks annotated for a limited set of seen classes. By training on seen annotations, models are encouraged to learn detailed features and patterns that are potentially applicable beyond the scope of the trained classes. 

Of most relevance, ZegFormer~\cite{ding2022decoupling} trains a MaskFormer by decoupling zero-shot semantic segmentation into two sub-tasks, a class-agnostic grouping task and a zero-shot segment classification task. Our method has similar training paradigm but with notable distinctions. Similar to GroupViT, we train exclusively with image-text pairs and do not utilize a pretrained CLIP model. Notably, even without access to ground truth masks, labels, or CLIP, our method outperforms ZegFormer in unseen categories, indicating potentially stronger generalization.

\emph{Custom grouping-based encoders.} The third strategy employs custom-designed models specifically for open-vocabulary segmentation. GroupViT~\cite{xu2022groupvit}  groups pixels in an image hierarchically based on their attention scores with learnable group tokens. OVSegmentor~\cite{xu2023learning} applies Slot Attention~\cite{locatello2020object} for a similar pixel grouping process based on feature proximity. 

Our model, \mname{}, can be conceptualized as a synergy of these approaches. It can be viewed as a CLIP model integrated with a MaskFormer image encoder, directly optimizing for pixel-level feature and language alignment. Alternatively, it resembles ``ZegFormer with pseudomask and language training'' or ``GroupVit with a MaskFormer as the grouping mechanism.'' Interestingly, our model relates to each method by omitting certain core architectural components or supervision method. Even so, our method is able to achieve competitive performance.

\textbf{Unsupervised image grouping.}
Unsupervised image grouping methods are designed to segment images without the use of manually labeled segmentation masks. 
Early unsupervised image grouping methods can be roughly categorized as low-level feature-based~\citep{canny1986computational}, clustering-based~\citep{kanungo2002efficient}, and graph-based~\citep{shi2000normalized}. 
More recently, self-supervised learning-based approaches~\citep{ji2019invariant, cho2021picie, hamilton2022unsupervised,van2020scan,van2021unsupervised,hwang2019segsort,zadaianchuk2023unsupervised} have shown superior performance in unsupervised image grouping.

\section{Approach}

Our proposed method, called \emph{\mname{}}, is conceptually simple: we learn a MaskFormer model from \emph{pseudo-mask} and \emph{language}. Our method leverages image-text pairs solely, without relying on ground truth masks or large-scale preatrained models. In figure~\ref{fig:pseudocode}, we provide pseudocode for the core implementation of training \mname{}. 
Figure~\ref{fig:main} provides a schematic layout of our approach.

\subsection{Problem definition}
\label{sec:prob_def}
We consider the problem of open-vocabulary semantic segmentation, where we aim to learn a function $f$ that maps an image $I$ and a set of category names $C=\{c_i\}$ to a semantic segmentation map $S$, where $c_i$ can be any category name expressed as open vocabulary texts.

Our approach is based on previous works \citep{xu2022groupvit, ranasinghe2022perceptual, zhou2022extract}, and we adopt their problem setting. Specifically, we use a web dataset of image-text pairs ${(I_i, T_i)}$ during training, where $T_i$ is a textual label that describes the content of the corresponding image $I_i$. However, since the textual labels are gathered from the web, they may be noisy and contain errors. We do not use any additional manual annotated segmentation or classification labels during training.

During testing, a set of category names $C$ is provided, and the model is tasked with assigning a semantic label $c_i\in C$ to each pixel in an unlabeled image. The performance of the model is evaluated based on its mean Intersection over Union (mIoU) with the ground truth labels.

\subsection{Adapting MaskFormer}
\label{sec:ov_maskfmr}
Our approach builds on top of MaskFormer~\citep{cheng2021per}. Here, we begin by briefly review MaskFormer and explain the adjustments we made. 

The Maskformer model
takes an image as input and generates $N$ masks and mask features. First, the input image passes through a backbone model to produce feature maps at different output resolutions. 
These image features are then fed into a per-pixel encoder, which upsamples and aggregates them into a set of feature maps with higher resolution. %
Meanwhile, a transformer decoder uses $N$ learnable queries to cross-attend to the set of features with the lowest resolution and gather global information about each segment.

In the original Maskformer, a linear classifier and softmax activation were applied to the output of the decoder to predict class probabilities for a fixed list of categories. However, as we do not have a fixed list of categories, we remove this classifier branch and output the $N$ raw mask features instead. 

In addition to predicting mask features, the Maskformer also predicts $N$ binary masks. 
To predict each mask, a dot product is taken between the mask embedding, generated from mask features, and the high resolution per-pixel feature. 
Finally, $N$ mask-feature pairs are combined to generate a semantic segmentation map as the output.

\subsection{\mname{}}
\label{sec:main_approach}
\mname{} employs MaskFormer as its segmentation model, but in our weakly-supervised learning setting (where only texts are available), we face the challenge of not having annotated masks and labels. To overcome this, we utilize pseudo labels and language to as supervision.

Our training framework is illustrated in Figure~\ref{fig:main}. We first generate a set of segmentation maps using our pseudo-mask generator (Sec.~\ref{sec:mask_sup}) and use them as supervision for mask prediction. Meanwhile, we use a language model to process input text and generate language embeddings. These embeddings provide supervision for mask features by leveraging image-text contrastive loss (Sec.~\ref{sec:sem_sup}).

\begin{figure}[t]
    \centering
    \includegraphics[width=.95\linewidth]{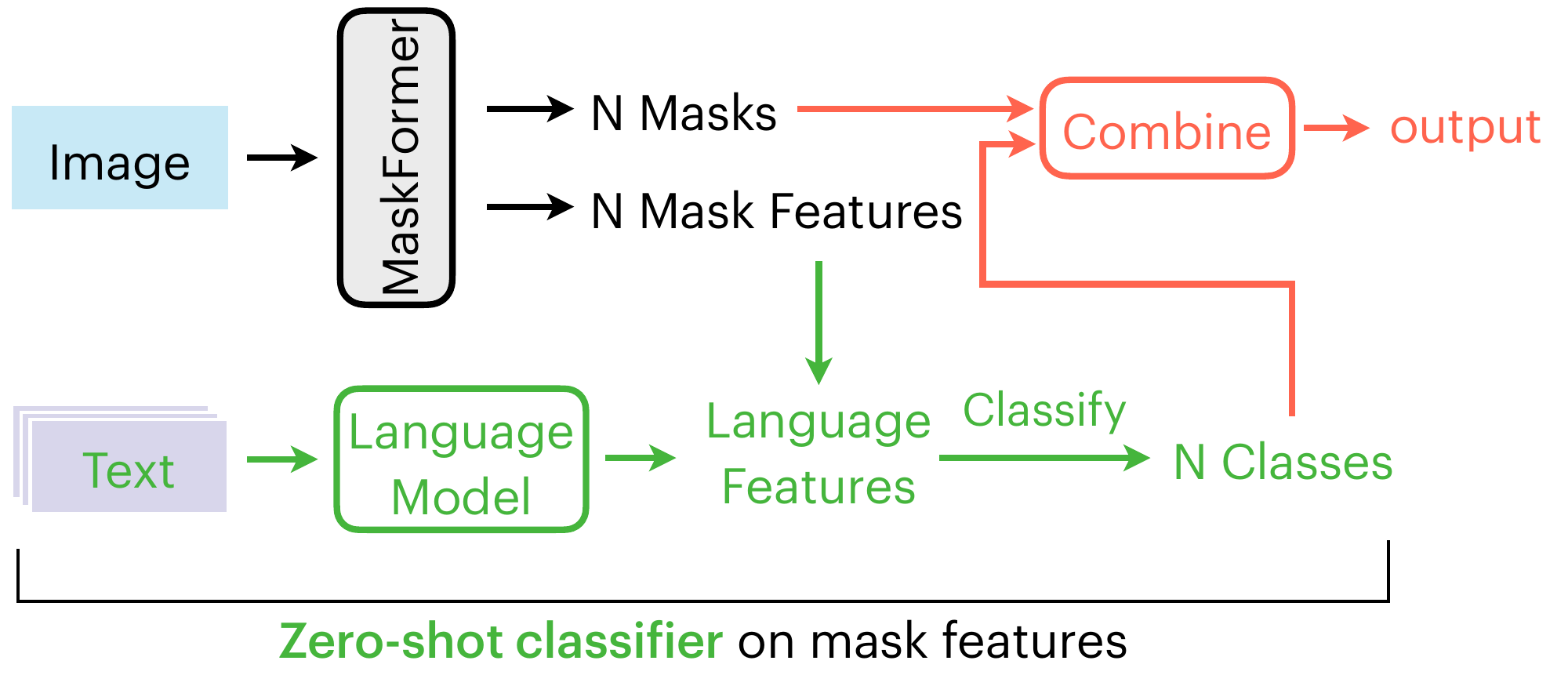}
    \caption{\textbf{Testing on \mname.} During inference, \mname{} generalize to new categories by leveraging language features generated from a list of candidate classes in text.}
    \vspace{-1em}
    \label{fig:dseg_test}
\end{figure}

Notably, unlike the supervised learning setting, where mask and label annotations are coupled, we \emph{decouple} mask and semantic supervision.  This enables us to utilize pseudo-mask and language as two distinct forms of supervision.

In the testing phase (as shown in figure~\ref{fig:dseg_test}), the trained MaskFormer model predicts $N$ masks and mask features from the input image. The language model takes as input a list of candidate category names (represented as texts) and extracts a set of language features. These features are then used to classify the mask features. This process is similar to the one used in CLIP~\citep{radford2021learning}, where the image and possible text inputs are encoded by their respective encoders to compute feature embeddings. The cosine similarity between these embeddings is calculated and adjusted by a learnable temperature parameter. The resulting values are normalized into a class probability distribution using a softmax function, and a combination module is used to takes $N$ mask-class pairs to produce the final segmentation map, similar to~\citep{cheng2021per}.

Next, we will provide a detailed description of the subcomponents in our framework.

\subsubsection{Pseudo-Mask Generator}
\label{sec:mask_sup}

\begin{figure}[t]
    \centering
    \includegraphics[width=0.95\linewidth]{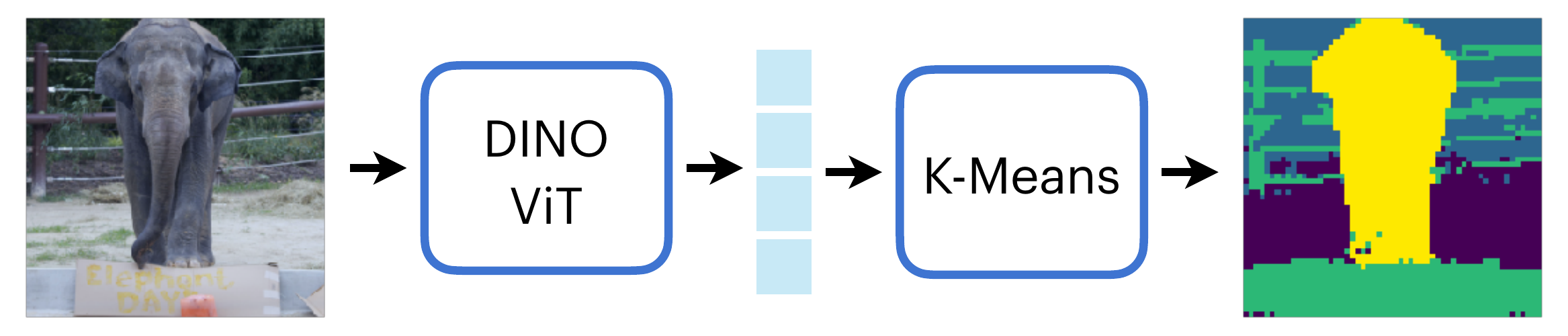}
    \caption{\textbf{The Pseudo-mask generator} generates pseudo-masks to supervise predicted mask during training. This module takes an image as its input, extracts its features using a DINO pre-trained ViT, and then employs K-means clustering to group the pixels into segments.}
    \vspace{-1em}
    \label{fig:pseudo_mask}
\end{figure}

\begin{figure}[t]
    \centering
    \includegraphics[width=0.95\linewidth]{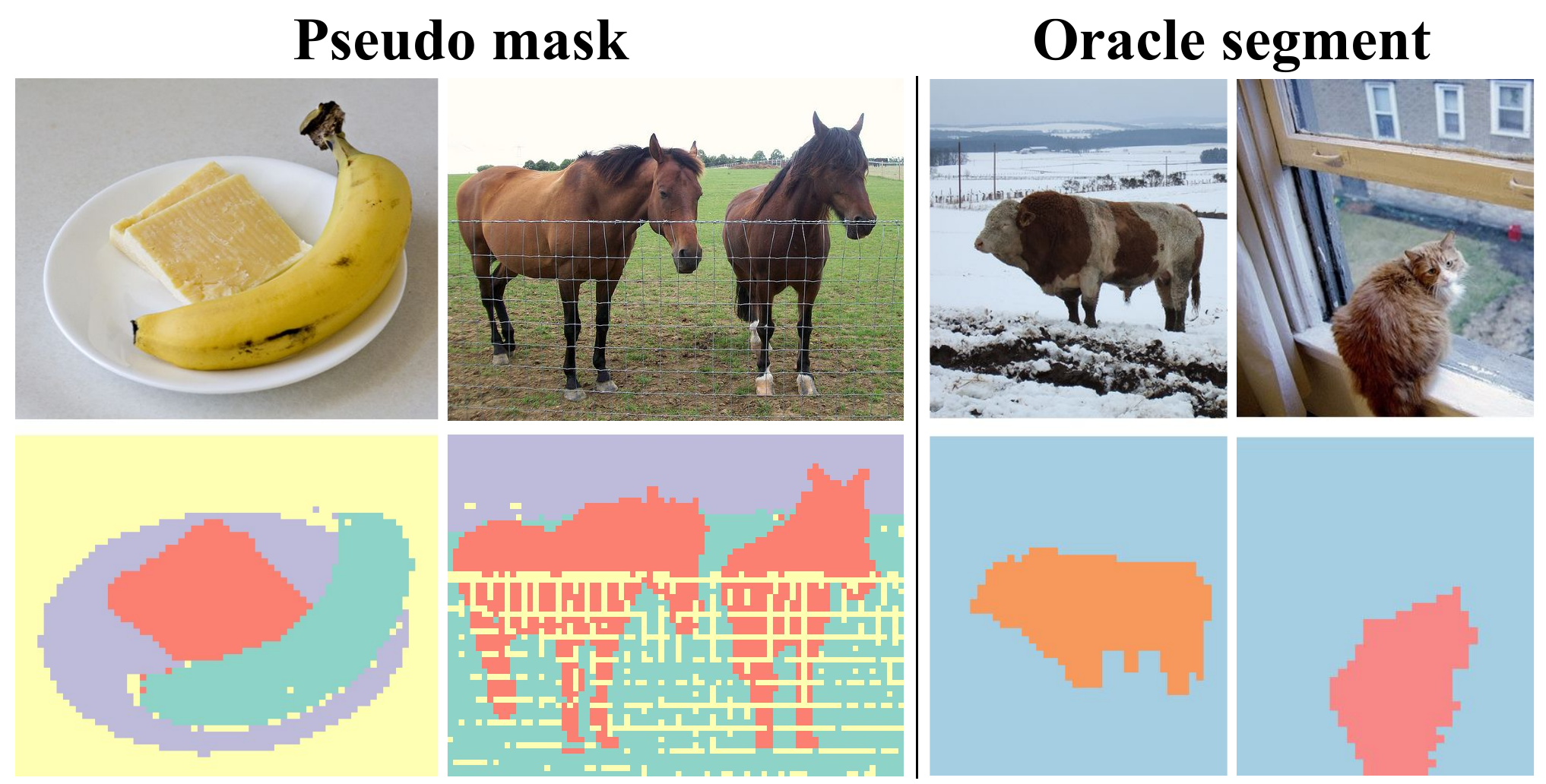}
\caption{\textbf{Example pseudo-masks.} Our pseudo-mask generator is capable of generating high-quality artificial masks. When provided with an oracle label, these masks demonstrate a high degree of overlap with the ground truth annotations.}
    \vspace{-1em}
    \label{fig:cluster}
\end{figure}

\begin{table}[t]
\centering
\tablestyle{5pt}{1.0}
\begin{tabular}{c|c|ccc}
Method  & Sup. &  P. VOC$\uparrow$ & P. Context$\uparrow$ & Time(s)$\downarrow$\\
\midrule
Spectral Clus.~\cite{shi2000normalized}* & none & 49.2 & 43.2 & 0.543 \\
K-Means~\cite{kanungo2002efficient}* & none & 49.5 & 43.3 & 0.188 \\
ImageNet~\cite{dosovitskiy2020image}+\cite{kanungo2002efficient}  & label & 68.8 & 58.1 & 0.079 \\
GroupViT~\cite{xu2022groupvit} & text &  73.7 & 54.6 & \textbf{0.002} \\
\midrule
\textbf{Pseudo-mask (Ours)} & self & \textbf{78.8} & \textbf{66.3} & \textbf{0.002} \\

\end{tabular}
\vspace{-0.05in}
\caption{\textbf{Our pseudo-mask generator achieves excellent oracle performance with rapid speed}, making it an ideal mask supervision. We report amortised running time on a batch of 128 samples, simulating training time scenario. * We process downsampled image at $\frac{H}{8}\times \frac{W}{8}$ resolution to obtain reasonable running time.}
\label{tab:oracle_perf}
\vspace{-1em}
\end{table}

In our approach, we use a pseudo-mask generator (fig.~\ref{fig:pseudo_mask}) to produce a class-agnostic segmentation map from the input image, which supervises the mask prediction of our model. 

To implement the pseudo-mask generator, we adopt a simple strategy that involves clustering tokens extracted from a self-supervised pre-trained ViT. Specifically, we use a DINO-pretrained ViT to compute a set of featurized tokens from the input image. We then apply a clustering algorithm (K-Means in our case) to these tokens, assigning each token a label that corresponds to the index of the cluster it belongs to. We reshape the resulting label map into an image and resize it to the original resolution to supervise the mask prediction of our segmentation model.

Despite its simplicity, our pseudo-mask generator achieves both impressive performance, which is crucial for high-quality supervision, and fast processing speed, which is essential for efficient training. We evaluate its performance and compare against baseline methods, and the quantitative results are presented in Table~\ref{tab:oracle_perf}, with example predictions visualized in~\ref{fig:cluster}. Our method significantly outperforms simple baselines such as K-Means and Spectral Clustering, which naively cluster image pixels, while running two orders of magnitude faster. We also observed that clustering DINO representation outperforms clustering ImageNet pre-trained ViT representation by a significant margin. Notably, our pseudo-mask generator even outperforms GroupViT, which has already employed vision-language training.

Since the predicted masks are unordered, we need to match the $N$ predicted masks with $K$ pseudo ground truth masks. To accomplish this, we utilize bipartite matching, as described in~\citep{carion2020end,cheng2021per}, which assigns a pseudo-mask to each predicted mask such that the overall assignment cost is minimal in all possible assignments. Since each pseudo-mask is assigned to at most one predicted mask, $N-K$ pseudo-masks are unassigned to no-object ($\O$). Unlike MaskFormer~\citep{cheng2021per}, we do not penalize these no-object masks, nor do we use classification loss as an assignment cost. Finally, we compute the mask loss between predicted masks and their corresponding pseudo-mask, utilizing a combination of dice loss~\citep{milletari2016v} and focal loss~\citep{lin2017focal}.

\begin{equation}
\mathcal{L}_{\text{mask}} = \lambda_{\text{dice}} \mathcal{L}_{\text{dice}} + \lambda_{\text{focal}} \mathcal{L}_{\text{focal}}
\end{equation}

\subsubsection{Language Supervision}
\label{sec:sem_sup}
Our model learns to classify open-vocabulary concepts from language supervision. To train the model, we use an image-text contrastive loss~\citep{radford2021learning, ghiasi2022scaling}. Specifically, we view $N$ mask features as representation of the input image, each capturing information about a different part of the image. We then compute a single feature that represents the entire image by taking the average of these mask features. To encode the text, we use a text transformer~\citep{vaswani2017attention} and select the embedding corresponding to the \texttt{[EOS]} token, resulting in a textual feature. Since the visual and textual features may have different dimensions, we project each representation into a common embedding space using 2-layer MLPs. To compute the image-text contrastive loss, we calculate the cosine similarity between the image embeddings and the text embeddings within the same batch. Following common practice~\citep{radford2021learning,mu2022slip,li2022scaling}, we decouple the image-text contrastive loss into two parts:

\begin{equation}
\mathcal{L}_{I\rightarrow T} = -\frac{1}{N}\sum_{i}^{N}\log\frac{\exp (x_i^\intercal y_i/\sigma)}{\sum^N_{j=1}\exp(x_i^\intercal y_j/\sigma)}
\end{equation}
\begin{equation}
\mathcal{L}_{T\rightarrow I} = -\frac{1}{N}\sum_{i}^{N}\log\frac{\exp (y_i^\intercal x_i/\sigma)}{\sum^N_{j=1}\exp(y_i^\intercal x_j/\sigma)}
\end{equation}

where $x_i$ and $y_i$ are L2-normalized embedding of image and text of the i-th pair. $N$ denotes batch size and $\sigma$ is a learnable temperature parameter optimized together with the rest of the model. The total loss is the sum of these two losses, $\mathcal{L}_{\text{contrastive}} = \mathcal{L}_{I\rightarrow T} + \mathcal{L}_{T\rightarrow I}$. This loss function promotes high similarity for positive pairs and low similarity for negative pairs. The loss is minimized when the positive image-text pairs have the highest similarity. To increase the contrastive efficiency, we aggregate negative samples 
from all nodes when we use distributed training, enabling more negative samples to be compared against.

\begin{figure*}[t]
    \centering
    \vspace{-1.5em}
    \includegraphics[width=0.99\linewidth]{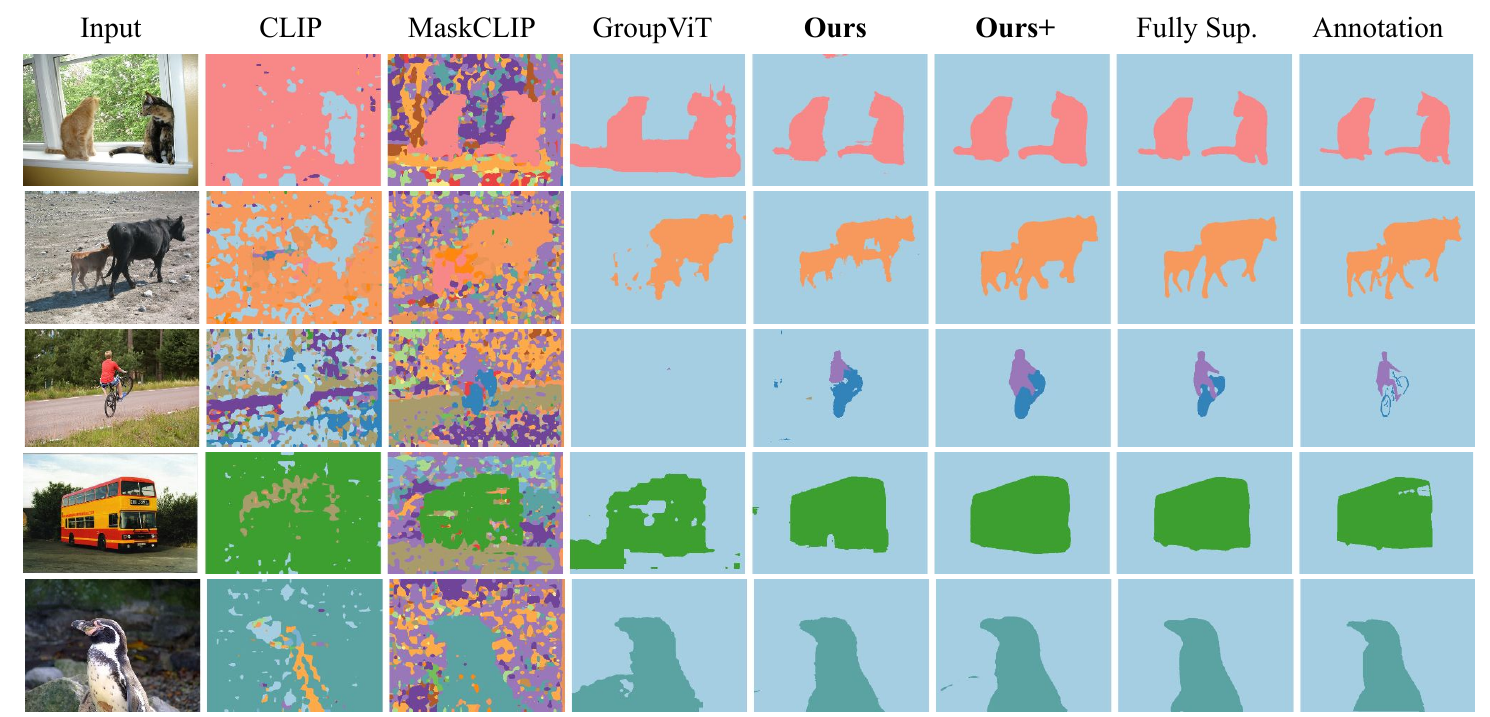}
    \vspace{-0.5em}
    \caption{\small \textbf{Qualitative comparison with existing methods.}  CLIP~\citep{radford2021learning} is primarily designed for classification and does not perform well in segmentation. MaskCLIP~\citep{zhou2022extract} adapts CLIP for segmentation, although it produces noisy predictions and cannot handle background classes. GroupViT~\citep{xu2022groupvit} is a strong competitor, but it could struggle in challenging scenarios. }
    \label{fig:main_compare}
\end{figure*}

\subsubsection{Training Loss}
\label{sec:loss}
Overall, mask loss (Sec.~\ref{sec:mask_sup}) and image-text contrastive loss (Sec.~\ref{sec:sem_sup}) complete the necessary mask and semantic supervision that is needed to train our model. The final loss is a weighted combination of the two losses:
\begin{equation}
L = \lambda_{\text{mask}} \mathcal{L}_{\text{mask}} + \lambda_{\text{contrastive}} \mathcal{L}_{\text{contrastive}}
\end{equation}

In our experiment, we use $\lambda_{\text{mask}}=1.0$, $\lambda_{\text{contrastive}}=1.0$, $\lambda_{\text{dice}}=1.0$, $\lambda_{\text{focal}}=20.0$.

\subsubsection{Self-training}
\label{sec:self_train}
In order to enhance our results, we introduce an optional step wherein we train a new model using the predictions generated by our current model. This process of self-training results in an augmented model, which we refer to as \mname{}+. More specifically, when we evaluate on a given dataset, we generate pseudo labels for the unlabeled images in the training set. Subsequently, we employ these pseudo labels to train a new segmentation model.

Self-training improves the accuracy by leveraging additional data~\citep{xie2020self}, augmentation~\citep{zoph2020rethinking}, and bootstrapping~\citep{grill2020bootstrap}. In our situation, self-training offers even greater benefits since we can take advantage of additional information that is obtainable during testing: unlabeled images and testing categories. We show that this additional step improves our results significantly at no extra manual labelling cost.

\section{Experiments}
In this section, we empirically evaluate our method and compare to existing approaches. We show that, although our method is quite simple, it performs surprisingly well against more complex existing methods. We evaluate the open-vocabulary semantic segmentation performance of \mname{} on the validation set of three datasets: Pascal VOC 2012~\citep{everingham2009pascal} (21 classes), Pascal Context~\citep{mottaghi2014role} (60 classes) and COCO~\citep{lin2014microsoft} (81 classes). For more implementation details, please refer to our supplementary materials.

\subsection{Simple baselines}
The high quality of pseudo-masks (as shown in Figure~\ref{fig:pseudo_mask}) may lead one to assume that the primary challenge is simply classifying these masks, and that this can be accomplished by utilizing pre-existing methods such as CLIP. To test this assumption, we first develop two simple baselines. 

\textbf{Baseline 1: Pseudo-mask + CLIP.} Firstly, our pseudo label generator is utilized to obtain pseudo segments. Then, we iterate through all the masks and apply the current mask to the original image. Next, the masked image is fed to CLIP for classification and the resulting class label is assigned to the corresponding segment.

\begin{table}[t]
\centering
\tablestyle{5pt}{1.0}
\begin{tabular}{c|cccc}
Method  & P. VOC & P. Context & COCO & 3-Avg.\\
\midrule
B1: Pseudo Mask + CLIP & 12.9 & 3.9 & 2.9 & 6.6 \\
B2: Pseudo-mask ViT & 23.2 & 11.0 & 10.4 & 14.9  \\
\midrule
\textbf{\mname{} (Ours)} &  \textbf{44.9} & \textbf{22.9} & \textbf{22.5} & \textbf{30.1} \\
\end{tabular}
\vspace{-0.05in}
\caption{\textbf{Simple baselines for open-vocabulary semantic segmentation.} All models are trained on CC12M. Higher values are better. Two simple baselines fail to obtain satisfactory results, even using after using our pseudo masks and no less training data.}
\label{tab:simple_baselines}
\vspace{-0.2in}
\end{table}

\textbf{Baseline 2: Pseudo-mask ViT.} We introduce a new visual backbone that differs from the regular ViT. Instead of pooling all image tokens into a single feature, we first individually pool tokens in each segment of the pseudo-mask into segment features, and then pool these features into a visual embedding. We train a CLIP-like model from scratch using this visual backbone. During testing, we classify each segment feature and assign the label to that segment.

\begin{table}[t]
\centering
\small
\tablestyle{5pt}{1.0}
\vspace{-1em}
\begin{tabular}{c|cc|ccc}
\toprule
Method & OV & Sup. & P. VOC & P. Context & COCO\\ 
\midrule
\multicolumn{6}{l}{\emph{Linearly-probed classification models:}}  \\
MoCo v3~\citep{chen2021empirical} & \xmark & self & 34.3 & 21.3 & - \\
DINO~\citep{caron2021emerging} & \xmark & self & 39.1 & 20.4 & - \\
\midrule
\multicolumn{6}{l}{\emph{Open-vocabulary models (annotated masks \underline{not} required for training):}} \\
CLIP~\citep{radford2021learning}$^\dagger$ & \cmark & text & 13.5 & 8.1 & 5.9 \\
MaskCLIP~\citep{zhou2022extract}$^\dagger$ & \cmark & text & 26.8 & 22.8 & 12.8  \\
ViL-Seg~\citep{liu2022open} & \cmark  & text & 34.4 & 16.3 & 16.4 \\
CLIP$_{\text{py}}$~\citep{ranasinghe2022perceptual} & \cmark & text & 52.2 & - & -  \\
GroupViT~\citep{xu2022groupvit} & \cmark  & text & 50.8 & 23.7 & 27.5  \\
SegCLIP~\citep{luo2023segclip} & \cmark  & text & 52.6 & 24.7 & 26.5  \\
OVSegmentor~\citep{xu2023learning} & \cmark  & text & 53.8 & 20.4 & 25.1  \\
TCL~\citep{cha2023learning} & \cmark  & text & \textbf{55.0} & \textbf{30.4} & \textbf{31.6}  \\
\midrule
\textbf{\mname{} (Ours)} & \cmark & text & {53.2} & {27.9} & {30.3} \\
\textbf{\mname{}+ (Ours)} & \cmark & text & \textbf{62.0}& \textbf{30.2} & \textbf{35.7} \\

\midrule
\multicolumn{6}{l}{\emph{Fully-supervised segmentation models:}} \\
DeepLabV3+$^\dagger$~\citep{chen2018encoder} & \xmark & GT & 78.7 & 46.4 & 55.7\\
MaskFormer$^\dagger$~\citep{cheng2021per} & \xmark & GT & 81.2 & 50.0 & 62.1\\
\bottomrule
\end{tabular}
\caption{\small \textbf{Open-vocabulary semantic segmentation results (background pixels included in evaluation).} Benchmarked on Pascal VOC (P. VOC), Pascal Context (P. Context) and COCO, following standard evaluation protocols on open-vocabulary model trained \emph{without} annotated masks~\cite{xu2022groupvit, luo2023segclip,xu2023learning,cha2023learning}. Our
approach obtain second highest performance on average and has better results than GroupViT on all datasets.
$^\dagger$ denotes our recomputed results. Higher values are better.}
\vspace{-1em}
\label{table:w_bg}
\end{table}

The results are presented in Table~\ref{tab:simple_baselines}. As we can see, open-vocabulary segmentation is more complex than simply  grouping image into segments and then categorizing them into classes,
even when the segments are of high quality. 
Baseline 1 employs a significantly larger pretrained CLIP ViT/L-14 model that was also trained on a much larger dataset, while Baseline 2 is trained using the same data as ours. Nevertheless, both baselines fail to achieve satisfactory results, suggesting that open-vocabulary segmentation cannot be naively deconstructed in such ways. We hypothesize that a multi-task learning approach that jointly trains for segmentation and classification could yield significant advantages.

\begin{table}[t]
\small
\centering
\vspace{-1em}
\tablestyle{4pt}{0.95}
\begin{tabular}{c |cc|cccc}
\toprule
Method & OV & Sup. &  P. VOC & P. Context & COCO\\ 
\midrule
\multicolumn{6}{l}{\emph{Open-vocabulary models (annotated masks required for training):}} \\
SPNet~\citep{xian2019semantic} & \cmark & mask+text & 18.3 & 24.3 & - \\
ZS3Net~\citep{bucher2019zero} & \cmark & mask+text & 38.3 & 19.4 & 21.1  \\
LSeg~\citep{li2022language} & \cmark & mask+text & 52.3 & - & 27.2 \\
OpenSeg~\citep{ghiasi2022scaling} & \cmark & mask+text & 77.2 & {45.9} & 38.1 \\
ZegFormer~\citep{ding2022decoupling} & \cmark & mask+text & 80.7 & - & - \\
GKC~\cite{han2023global} & \cmark & mask+text & 83.2 & 45.2 & - \\
ODISE~\cite{xu2023open} & \cmark & mask+text & 85.7 & 84.6 & \textcolor{gray} {\textbf{65.2}*} \\
DeOp~\cite{han2023open} & \cmark & mask+text & 91.7 & 48.8 & - \\
OVSeg~\cite{liang2023open} & \cmark & mask+text & 94.5 & 55.7 & - \\
SAN~\cite{xu2023side} & \cmark & mask+text & \textbf{94.6} & \textbf{57.7} & - \\
\midrule
\multicolumn{6}{l}{\emph{Open-vocabulary models (annotated masks \underline{not} required for training):}} \\
CLIP~\citep{radford2021learning}$^\dagger$ & \cmark & text & 39.6 & 	9.0 & 13.8\\
MaskCLIP~\citep{zhou2022extract}$^\dagger$ & \cmark & text & 49.5	& 25.5 & 23.6  \\
GroupViT~\citep{xu2022groupvit}$^\dagger$ & \cmark & text & 77.2 & 23.0 & 37.5 \\
TCL~\citep{cha2023learning} & \cmark & text & \textbf{83.2} & \textbf{33.9} & - \\
\midrule
\textbf{\mname{}  (Ours)} & \cmark & text & 
{81.8}& 
{27.2}&
{42.4}\\
\textbf{\mname+ (Ours)}& \cmark & text & 
\textbf{84.7}& 
\textbf{31.6}& 
\textbf{53.0}\\

\midrule
\multicolumn{6}{l}{\emph{Fully-supervised segmentation models:}} \\
DeepLabV3+$^\dagger$~\citep{chen2018encoder}& \xmark & GT & 89.9  & 48.5 & 66.9 \\
\bottomrule
\end{tabular}
\caption{\small \textbf{Open-vocabulary semantic segmentation results (background pixels excluded in evaluation).} Benchmarked following standard protocol for evaluating open-vocabulary models \emph{with} annotated masks~\cite{han2023global,ding2022decoupling, han2023open}. \mname{} achieves competitive performance compared to earlier methods similar to the previous setting.
 $^\dagger$ denotes our recomputed results. *COCO is used for training. Higher values are better. } 
\vspace{-1em}
\label{table:main}
\end{table}

\begin{figure*}[t]
    \vspace{-0.2em}
    \centering
    \subfloat[\centering Pascal VOC ($+18.3\%$) ]{{\includegraphics[width=0.3\linewidth]{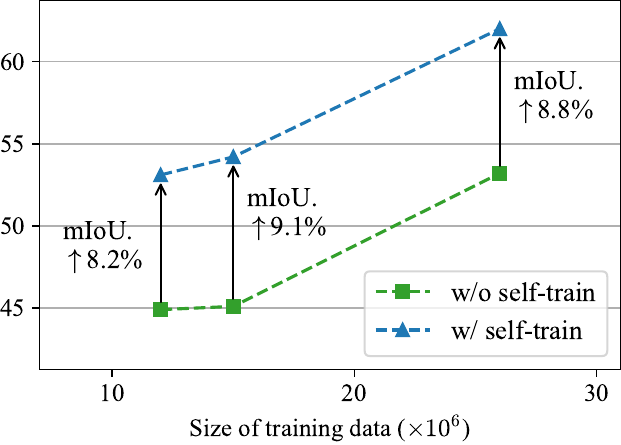} }}%
    \ 
    \subfloat[\centering Pascal Context ($+14.1\%$)]{{\includegraphics[width=0.3\linewidth]{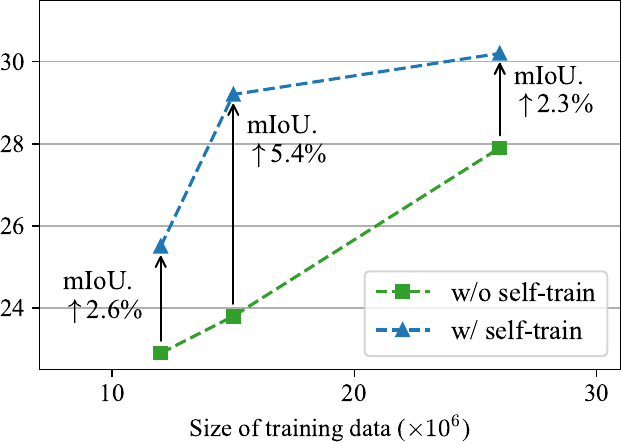} }}%
    \ 
    \subfloat[\centering COCO ($+17.6\%$)]{{\includegraphics[width=0.3\linewidth]{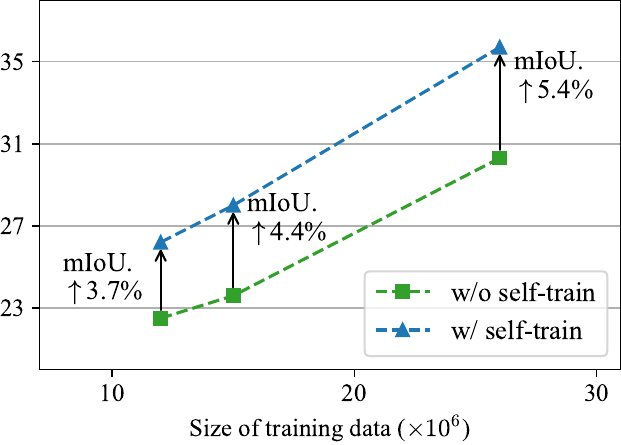} }}%
    \vspace{-0.5em}
    \caption{\small \textbf{Self-training improvement.} We show average relative improvement in bracket on top of the plot. we observe that self-training consistently leads to significant improvement for \mname{} across all of our training and testing data settings. }
    \label{fig:self-train}
\end{figure*}

\subsection{Evaluation with background}
In table~\ref{table:w_bg}, we evaluate our model and compare with existing method on open-vocabulary semantic segmentation task. Following standard evaluation protocols on open-vocabulary model trained \emph{without} annotated masks~\cite{xu2022groupvit, luo2023segclip,xu2023learning,cha2023learning}, we include background pixels in evaluation and obtain background prediction by setting a theshold for background classes~\cite{xu2022groupvit}. Despite the simplicity of \mname{}, our approach achieve competitive performance over previous open-vocabulary segmentation methods that does not require mask annotations. Our model has second highest performance on average and has better results than GroupViT on all datasets. Moreover, our self-trained model, \mname{}+, provides an impressive 5.5\% mIoU improvement over our base model \mname{} (42.6\% vs 37.1\% 3-avg. mIoU), suggesting the efficacy of self-training.

\subsection{Evaluation without background}
We also evaluate our model on the evaluation protocol commonly used for evaluating open-vocabulary models \emph{with} annotated masks~\cite{han2023global,ding2022decoupling, han2023open}, where the background  pixels are excluded in evaluation. We note that this setting is easier because background class is more diverse in appearance and often requires additional processing such as thresholding. Table~\ref{table:main} shows the results. Similar to the previous setting, our \mname{} and \mname{}+ models achieve competitive performance compared to earlier methods.

\subsection{Ablation studies}
\label{sec:ablation}

\textbf{Self-training.} We investigated the effectiveness of self-training for improving segmentation performance. To this end, we compared \mname{} and \mname{}+ on three datasets and evaluated the results using Figure~\ref{fig:self-train}. We found that self-training consistently improved the segmentation performance by a significant margin ($+5.5\%$ mIoU on average), regardless of the data size and test dataset. These results indicate that self-training is a reliable approach for enhancing the performance of \mname{} and can provide a desirable complement for further improvement.

\textbf{Data scalability.}
To evaluate the scalability of our method, we trained \mname{} and \mname{}+ using three datasets of increasing sizes: 12M, 15M, and 26M. The results of the experiments are presented in 
Figure~\ref{fig:data_scale}. We observed that both models achieve significant improvements in performance across all three testing datasets as the amount of data increased, suggesting that our method scales well with larger datasets.

\begin{figure}[t]
    \centering
    \includegraphics[width=\linewidth]{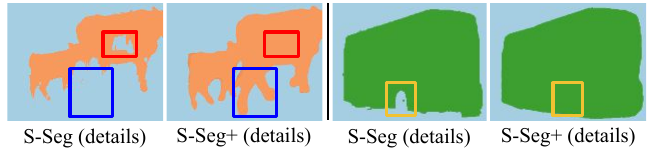}
    \vspace{-1.6em}
    \caption{ \textbf{Visualizing effect of self-training.} Our self-trained \mname+ model demonstrates the ability to accurately predict in regions overlooked by \mname, as shown in the  colorful rectangles.}
    \label{fig:abl_st}
\end{figure}

\begin{figure}[t]
    \centering
        \vspace{-0.5em}
    \subfloat[\centering \mname{}]{{\includegraphics[width=0.95\linewidth]{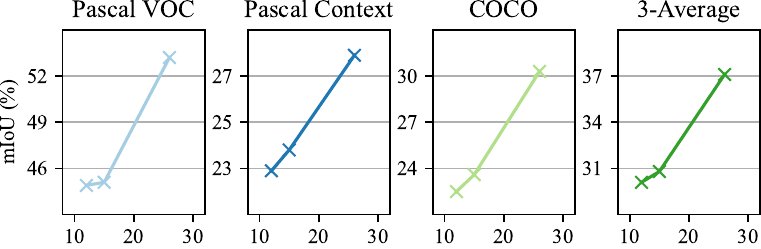} }}%
    
    \vspace{-0.5em}
    
    \subfloat[\centering \mname+]{{\includegraphics[width=0.95\linewidth]{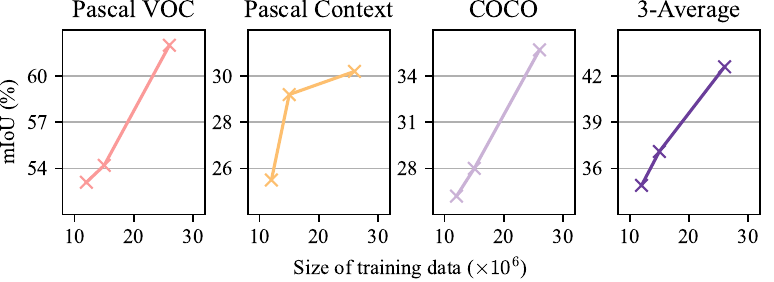} }}%
    \vspace{-0.2em}

    \caption{ \textbf{Scaling training data provide consistent gain in performance, with or without self-training.} We train our model using different sizes of data: CC12M (12M), CC12M+CC3M (15M), and CC12M+CC3M+RedCaps (26M). We note a steady improvement in the performance as the data size increases.}

    \vspace{-1em}
    \label{fig:data_scale}
\end{figure}

\subsection{Visualization}

\begin{figure}[t]
    \centering
    \includegraphics[width=0.83\linewidth]{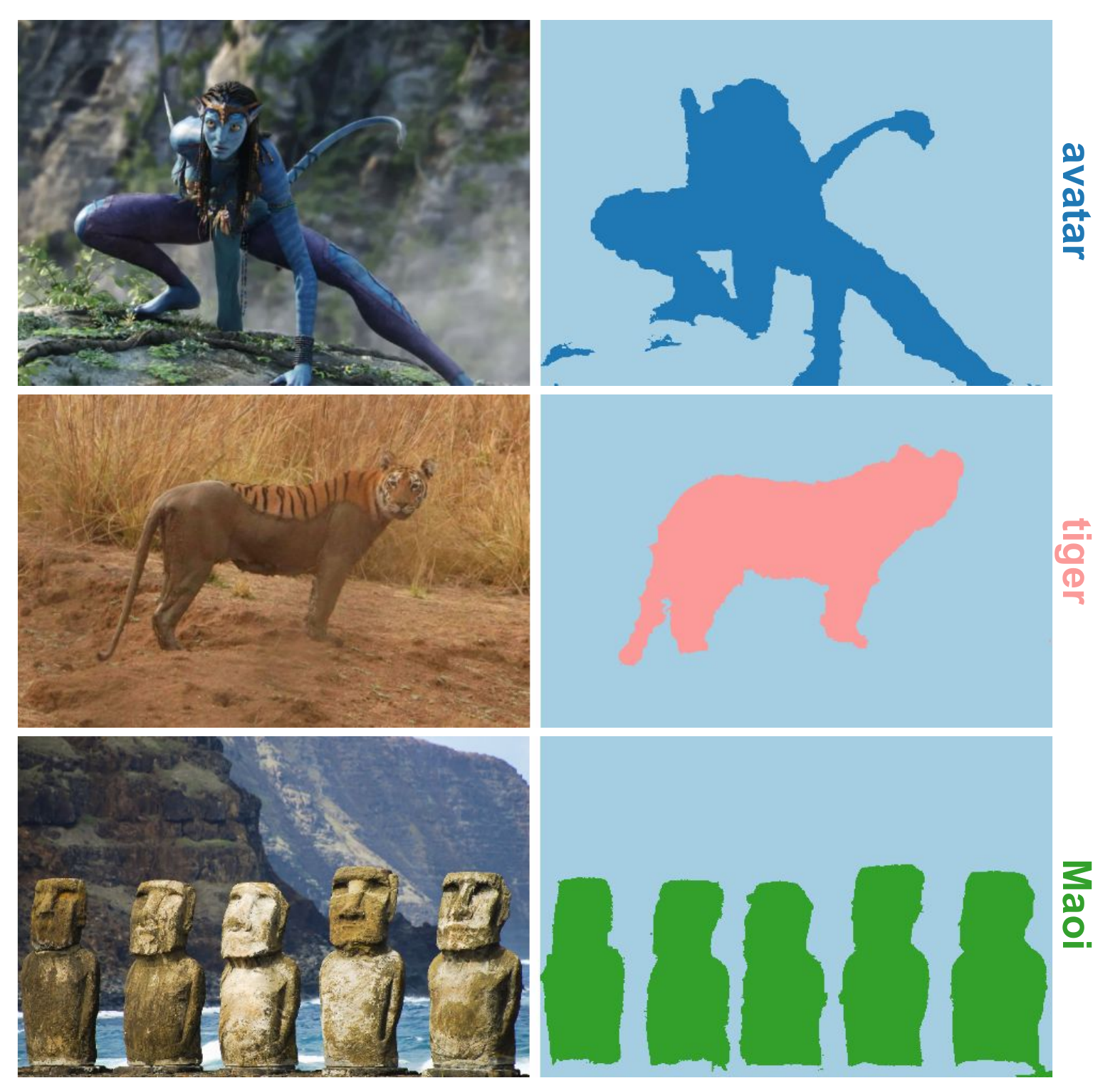}
        \vspace{-0.3em}
    \caption{\textbf{Qualitative results on web images.} The query class name is shown to the right. Row 1: \mname{} is able to segment fictional characters in an animated scene.
Row 2: Despite having taken a mud bath, the tiger can still be easily recognized and segmented.
Row 3: \mname{} is capable of identify specific landmarks.}
    \vspace{-1.3em}
    \label{fig:web_image}
\end{figure}

The qualitative results of our model are illustrated in Figure~\ref{fig:qualitative}. Our model has demonstrated its ability to handle difficult situations such as overlapping and small objects. Comparing our results to those of existing methods, as shown in Figure~\ref{fig:main_compare}, we observed that our approach accurately segments objects in challenging cases where previous methods have failed. 
Additionally, we observed that self-training can correct minor errors in our base model (as shown in detail in fig.~\ref{fig:abl_st}). In Figure~\ref{fig:teaser} and \ref{fig:web_image}, we present \mname{}'s performance on web images using custom query classes. Our model is able to produce precise results for these categories. For more qualitative results, please refer to our supplementary material.

\section{Conclusion}
To summarize, we propose \mname{}, a simple and intuitive framework that enables accurate and generalizable open-vocabulary segmentation. Our algorithm directly trains for pixel-level feature and language alignment, and does not require manual segmentation annotations or extensive pretraining. We hope that
our simple yet effective approach will serve as a solid baseline for future research.

{
    \small
    \bibliographystyle{ieeenat_fullname}
    \bibliography{shortstrings,cvpr}
}

\clearpage
\appendix

\section{Implementation details}
\label{sec:implementation}
\subsection{\mname{} experiments}
\textbf{Architecture.}
Our experiments use MaskFormer~\cite{cheng2021per} with Swin-S~\cite{liu2021swin} backbone and 6-layer transformer decoder with $N=64$ queries. The hidden and output feature dimension is 256. The language model is a Transformer~\cite{vaswani2017attention} with 12 layers, each with a hidden dimension of 256.  The context length (maximum length of input text) is set to 77 and the vocabulary size is 49408. We use a 2-layer MLP to project the visual and text feature into a common embedding space of dimension 256. We use DINO ViT-S/8 as the pretrained ViT in pseudo-mask generator and generates 8 pseudo-masks. 

\textbf{Training.}
\label{sec:train_details}
During training, we used three publically available datasets: CC3M~\cite{sharma2018conceptual}, CC12M~\cite{changpinyo2021conceptual}, and RedCaps~\cite{desai2021redcaps}, containing 3M, 12M and 12M image-text pairs, respectively. 
Due to storage constraint, we use only first 11M data samples at a smaller resolution of 
when using RedCaps dataset. In total, we use at most 26M image-text pairs for training - this is an order of magnitude fewer data than CLIP~\cite{radford2021learning} and 1-4M fewer than GroupViT~\cite{xu2022groupvit}. 
The total dataset takes about 2.4 TB storage space. 
Table~\ref{tab:impl_pseg} shows our default training setting.
All input images are random resized and cropped to $224\times 224$ in resolution. Following~\cite{xu2022groupvit}, we extract nouns and verbs from raw sentence because these words are more likely to describe the image.

\textbf{Inference.} We evaluate \mname{} on the validation set of three datasets: Pascal VOC 2012~\cite{everingham2009pascal}, Pascal Context~\cite{mottaghi2014role} and COCO~\cite{lin2014microsoft}. 
The Pascal VOC dataset contain 1449 images for testing. Each image is labeled with 20 foreground classes and a background class. The Pascal Context dataset contains 5104 testing images with 59 foreground classes and a background class. The COCO dataset contains 5000 images for testing with 80 foreground classes and an additional background class. 
As in~\cite{xu2022groupvit}, we combine all instances of the same class to get semantic segmentation mask for each image in COCO. Following GroupViT~\cite{xu2022groupvit}, we threshold the maximum probability to obtain background prediction. 
 During inference, we set the input resolution to $448 \times 448$, which is consistent with~\cite{xu2022groupvit}.

\subsection{\mname{}+ experiments}
\textbf{Self-training.} For self-training experiments, we use UperNet~\cite{xiao2018unified} with MAE~\cite{he2022masked} pretrained ViT backbone. We utilize a pyramid-structured network to merge the features obtained from layer 4, 6, 8, and 12 of the ViT, following the implementation of BEiT~\cite{bao2021beit}. We use the same model that we used to evaluate our main results to generate training data from the train set of the respective dataset. Training hyperparameters are provided in Table~\ref{tab:impl_pseg_plus}. Following~\cite{bao2021beit,he2022masked}, we use a layerwise learning rate decay~\cite{clark2020electra}. We do \emph{not} use relative position embeddings in our backbone ViT model (which is used by~\cite{bao2021beit,he2022masked} at fine-tuning stage for extra improvement).

\begin{table}[t]
\tablestyle{6pt}{1.02}
\begin{tabular}{y{86}|y{108}}
config & value \\
\shline
optimizer & AdamW \cite{loshchilov2017decoupled} \\
base learning rate & 5e-4 \\
weight decay & 0.05 \\
optimizer momentum & $\beta_1, \beta_2{=}0.9, 0.999$\\
batch size & 4096 \\
learning rate schedule & cosine decay \cite{loshchilov2016sgdr} \\
warmup epochs \cite{goyal2017accurate} & 2 \\
training epochs & 30 \\
\end{tabular}
\vspace{.5em}
\caption{\textbf{\mname{} setting.}}
\label{tab:impl_pseg} %
\end{table}

\begin{table}[t]
\tablestyle{6pt}{1.02}
\begin{tabular}{y{86}|y{108}}
config & value \\
\shline
optimizer & AdamW \cite{loshchilov2017decoupled} \\
base learning rate & 1e-4 \\
weight decay & 0.05 \\
optimizer momentum & $\beta_1, \beta_2{=}0.9, 0.999$\\
batch size & 16 \\
learning rate schedule & polynomial decay \\
warmup iters \cite{goyal2017accurate} & 1.5k \\
training iters & 20k (voc), 40k (ctxt), 80k (coco) \\
layer-wise lr decay~\cite{clark2020electra} & 0.7 \\
\end{tabular}
\vspace{.5em}
\caption{\textbf{\mname{}+ setting.}}
\label{tab:impl_pseg_plus} \vspace{-.5em}
\end{table}

\begin{table*}[t]\centering 
\subfloat[\textbf{Scaling training data provide consistent gain}:  We train our model using different size of data: 12M (CC12M), 15M (+CC3M), and 26M (+RedCaps). We note a steady improvement in the model's performance as the data size increases. \label{tab:ablation:training_data}]{
\tablestyle{4pt}{1.1}
\begin{tabular}{x{22}|x{33}x{33}x{33}|x{33}x{33}x{33}}
\multirow{2}{*}{data} & \multicolumn{3}{c|}{\textbf{\mname}} & \multicolumn{3}{c}{\textbf{\mname+}}\\
& VOC & Context & COCO & VOC & Context & COCO \\
\shline
12M & 44.9 & 22.9 & 22.5 & 53.1 & 25.5 & 26.2\\
15M & 45.1\textcolor{blue}{\scriptsize (+0.2)} & 23.8\textcolor{blue}{\scriptsize (+0.9)} & 27.9\textcolor{blue}{\scriptsize (+5.4)} & 54.2\textcolor{blue}{\scriptsize (+1.1)} & 29.2\textcolor{blue}{\scriptsize (+3.7)} & 28.0\textcolor{blue}{\scriptsize (+1.8)} \\
26M & {\textbf{53.2}\textcolor{blue}{\scriptsize (+8.3)}} & {\textbf{27.9}}\textcolor{blue}{\scriptsize (+5.0)} & {\textbf{30.3}}\textcolor{blue}{\scriptsize (+7.8)} & {\textbf{62.0}}\textcolor{blue}{\scriptsize (+8.9)} & {\textbf{30.2}}\textcolor{blue}{\scriptsize (+4.7)} & {\textbf{35.7}}\textcolor{blue}{\scriptsize (+9.5)}  \\
\end{tabular}}\hspace{3mm}
\subfloat[\textbf{Self-training offers constant improvement}: We observe that self-training consistently leads to significant improvement on performance across 3 datasets. \label{tab:ablation:mask_inference}]{
\tablestyle{4pt}{1.1}
\begin{tabular}{x{52}|x{32}x{32}x{32}}
\multirow{2}{*}{method} & \multicolumn{3}{c}{\textbf{3-Average}}\\
  & 12M & 15M & 26M \\
\shline
w/o self-train  &  30.1 & 30.8 & 37.1  \\
w/ self-train & \textbf{34.9} & \textbf{37.1} & \textbf{42.6}  \\
\hline
$\Delta$ &  \textcolor{blue}{+4.8} & \textcolor{blue}{+6.3} & \textcolor{blue}{+5.5} \\
 
\end{tabular}}\hspace{3mm}

\caption{\textbf{Ablations on data scalability and self-training}. We report mIoU evaluated on three datasets. Higher values are better.}
\label{tab:ablations_supp}
\end{table*}

\begin{table*}[t]
    \linespread{1}
    \aboverulesep = 0.2em \belowrulesep = 0.2em
    \scriptsize
    \centering
    \begin{tabular}{cc|ccccccccccccccccccccccc} 
        && \hspace{-1.2em} & \rotatebox[origin=lb]{90}{\smash{BG.}} & \rotatebox[origin=lb]{90}{\smash{aeroplane}} & \rotatebox[origin=lb]{90}{\smash{bicycle}} & \rotatebox[origin=lb]{90}{\smash{bird}} & \rotatebox[origin=lb]{90}{\smash{boat}} & \rotatebox[origin=lb]{90}{\smash{bottle}} & \rotatebox[origin=lb]{90}{\smash{bus}} & \rotatebox[origin=lb]{90}{\smash{car}} & \rotatebox[origin=lb]{90}{\smash{cat}} & \rotatebox[origin=lb]{90}{\smash{chait}} & \rotatebox[origin=lb]{90}{\smash{cow}} & \rotatebox[origin=lb]{90}{\smash{table}} & \rotatebox[origin=lb]{90}{\smash{dog}} & \rotatebox[origin=lb]{90}{\smash{horse}} & \rotatebox[origin=lb]{90}{\smash{motorbike}} & \rotatebox[origin=lb]{90}{\smash{person}} & \rotatebox[origin=lb]{90}{\smash{plant}} & \rotatebox[origin=lb]{90}{\smash{sheep}} & \rotatebox[origin=lb]{90}{\smash{sofa}} & \rotatebox[origin=lb]{90}{\smash{train}} & \rotatebox[origin=lb]{90}{\smash{monitor}}& \rotatebox[origin=lb]{90}{\smash{mIoU}}
        \\
        \midrule
        \multirow{5}{0em}{\rotatebox[origin=c]{90}{OV Methods}}
& CLIP & \hspace{-1.2em} & 13.2\hspace{-0.3em}&\hspace{-0.8em}10.4\hspace{-0.3em}&\hspace{-0.8em}4.4\hspace{-0.3em}&\hspace{-0.8em}8.0\hspace{-0.3em}&\hspace{-0.8em}5.9\hspace{-0.3em}&\hspace{-0.8em}19.4\hspace{-0.3em}&\hspace{-0.8em}27.0\hspace{-0.3em}&\hspace{-0.8em}17.5\hspace{-0.3em}&\hspace{-0.8em}26.0\hspace{-0.3em}&\hspace{-0.8em}3.1\hspace{-0.3em}&\hspace{-0.8em}19.6\hspace{-0.3em}&\hspace{-0.8em}9.0\hspace{-0.3em}&\hspace{-0.8em}21.5\hspace{-0.3em}&\hspace{-0.8em}16.8\hspace{-0.3em}&\hspace{-0.8em}11.2\hspace{-0.3em}&\hspace{-0.8em}11.7\hspace{-0.3em}&\hspace{-0.8em}5.2\hspace{-0.3em}&\hspace{-0.8em}13.1\hspace{-0.3em}&\hspace{-0.8em}7.6\hspace{-0.3em}&\hspace{-0.8em}21.1\hspace{-0.3em}&\hspace{-0.8em}12.2\hspace{-0.3em}&\hspace{-0.8em}13.5\hspace{-0.3em}\\
& MaskCLIP & \hspace{-1.2em} & 41.3\hspace{-0.3em}&\hspace{-0.8em}12.8\hspace{-0.3em}&\hspace{-0.8em}18.7\hspace{-0.3em}&\hspace{-0.8em}22.5\hspace{-0.3em}&\hspace{-0.8em}6.7\hspace{-0.3em}&\hspace{-0.8em}22.8\hspace{-0.3em}&\hspace{-0.8em}50.7\hspace{-0.3em}&\hspace{-0.8em}23.4\hspace{-0.3em}&\hspace{-0.8em}56.8\hspace{-0.3em}&\hspace{-0.8em}13.6\hspace{-0.3em}&\hspace{-0.8em}34.1\hspace{-0.3em}&\hspace{-0.8em}8.1\hspace{-0.3em}&\hspace{-0.8em}46.3\hspace{-0.3em}&\hspace{-0.8em}29.5\hspace{-0.3em}&\hspace{-0.8em}39.9\hspace{-0.3em}&\hspace{-0.8em}22.7\hspace{-0.3em}&\hspace{-0.8em}9.5\hspace{-0.3em}&\hspace{-0.8em}29.5\hspace{-0.3em}&\hspace{-0.8em}25.1\hspace{-0.3em}&\hspace{-0.8em}30.8\hspace{-0.3em}&\hspace{-0.8em}18.2\hspace{-0.3em}&\hspace{-0.8em}26.8\hspace{-0.3em}\\
& GroupViT & \hspace{-1.2em} & 79.0\hspace{-0.3em}&\hspace{-0.8em}37.4\hspace{-0.3em}&\hspace{-0.8em}29.9\hspace{-0.3em}&\hspace{-0.8em}33.3\hspace{-0.3em}&\hspace{-0.8em}33.9\hspace{-0.3em}&\hspace{-0.8em}64.4\hspace{-0.3em}&\hspace{-0.8em}60.2\hspace{-0.3em}&\hspace{-0.8em}62.4\hspace{-0.3em}&\hspace{-0.8em}76.7\hspace{-0.3em}&\hspace{-0.8em}16.2\hspace{-0.3em}&\hspace{-0.8em}68.8\hspace{-0.3em}&\hspace{-0.8em}28.0\hspace{-0.3em}&\hspace{-0.8em}75.9\hspace{-0.3em}&\hspace{-0.8em}62.5\hspace{-0.3em}&\hspace{-0.8em}64.2\hspace{-0.3em}&\hspace{-0.8em}51.6\hspace{-0.3em}&\hspace{-0.8em}38.7\hspace{-0.3em}&\hspace{-0.8em}63.0\hspace{-0.3em}&\hspace{-0.8em}37.4\hspace{-0.3em}&\hspace{-0.8em}44.0\hspace{-0.3em}&\hspace{-0.8em}38.4\hspace{-0.3em}&\hspace{-0.8em}50.8\hspace{-0.3em}\\
& \textbf{\mname{}(Ours)} & \hspace{-1.2em} & 81.0\hspace{-0.3em}&\hspace{-0.8em}47.2\hspace{-0.3em}&\hspace{-0.8em}40.1\hspace{-0.3em}&\hspace{-0.8em}38.6\hspace{-0.3em}&\hspace{-0.8em}30.0\hspace{-0.3em}&\hspace{-0.8em}63.5\hspace{-0.3em}&\hspace{-0.8em}74.6\hspace{-0.3em}&\hspace{-0.8em}67.6\hspace{-0.3em}&\hspace{-0.8em}75.7\hspace{-0.3em}&\hspace{-0.8em}\textbf{18.6}\hspace{-0.3em}&\hspace{-0.8em}65.3\hspace{-0.3em}&\hspace{-0.8em}34.4\hspace{-0.3em}&\hspace{-0.8em}72.2\hspace{-0.3em}&\hspace{-0.8em}56.3\hspace{-0.3em}&\hspace{-0.8em}68.0\hspace{-0.3em}&\hspace{-0.8em}50.7\hspace{-0.3em}&\hspace{-0.8em}45.7\hspace{-0.3em}&\hspace{-0.8em}60.2\hspace{-0.3em}&\hspace{-0.8em}33.6\hspace{-0.3em}&\hspace{-0.8em}53.1\hspace{-0.3em}&\hspace{-0.8em}41.0\hspace{-0.3em}&\hspace{-0.8em}53.2\hspace{-0.3em}\\
& \textbf{\mname+ (Ours)} & \hspace{-1.2em} & \textbf{86.5}\hspace{-0.3em}&\hspace{-0.8em}\textbf{53.8}\hspace{-0.3em}&\hspace{-0.8em}\textbf{42.0}\hspace{-0.3em}&\hspace{-0.8em}\textbf{48.1}\hspace{-0.3em}&\hspace{-0.8em}\textbf{49.3}\hspace{-0.3em}&\hspace{-0.8em}\textbf{76.0}\hspace{-0.3em}&\hspace{-0.8em}\textbf{84.7}\hspace{-0.3em}&\hspace{-0.8em}\textbf{74.5}\hspace{-0.3em}&\hspace{-0.8em}\textbf{87.2}\hspace{-0.3em}&\hspace{-0.8em}{17.1}\hspace{-0.3em}&\hspace{-0.8em}\textbf{81.8}\hspace{-0.3em}&\hspace{-0.8em}\textbf{35.0}\hspace{-0.3em}&\hspace{-0.8em}\textbf{83.4}\hspace{-0.3em}&\hspace{-0.8em}\textbf{65.2}\hspace{-0.3em}&\hspace{-0.8em}\textbf{74.3}\hspace{-0.3em}&\hspace{-0.8em}\textbf{65.3}\hspace{-0.3em}&\hspace{-0.8em}\textbf{46.6}\hspace{-0.3em}&\hspace{-0.8em}\textbf{78.2}\hspace{-0.3em}&\hspace{-0.8em}\textbf{40.2}\hspace{-0.3em}&\hspace{-0.8em}\textbf{58.5}\hspace{-0.3em}&\hspace{-0.8em}\textbf{53.6}\hspace{-0.3em}&\hspace{-0.8em}\textbf{62.0}\hspace{-0.3em}\\
\bottomrule
    \end{tabular}
    \caption{\textbf{Per-category open vocabulary semantic segmentation performance over 21 Pascal VOC classes.} Our method surpass baseline methods such as GroupViT on the Pascal VOC dataset, particularly in segmenting large objects and categories with consistent textures.}
    \label{tab:voc_per_class}
\end{table*}

\subsection{Reimplemented baselines}
\textbf{CLIP~\cite{radford2021learning}.} We utilized the CLIP ViT-B/16 model along with the official pretraining weights. The ViT model incorporates attentional pooling in its last layer, using an additional [CLS] token to aggregate other tokens. We choose to employ the \emph{value} embedding as the representation of each token, as the query and key embedding of the final layer is not fully trained during CLIP pretraining (only the similarity between the query embedding of the [CLS] token and the key embedding of other tokens is utilized). Finally, we leverage the language model to encode all classes and classify the visual tokens, similar to CLIP's zero-shot classification approach.

\textbf{MaskCLIP~\cite{cheng2021per}.} We use the testing code and weights provided by the authors, but re-evaluating them on the commonly-used protocol that includes the background class. To further assess the efficacy of our approach, as well as baseline methods, we employed the evaluation metric utilized by MaskCLIP, which specifically disregards background pixels.

\textbf{GroupViT~\cite{xu2022groupvit}.} The GroupViT project has provided pre-trained models for two configurations. Without specific clarification, we opt to use the model with the highest average accuracy, which was trained on CC12M, CC15M, and Redcaps datasets. This particular model also closely aligns with our method in terms of training data. 

\textbf{Fully supervised models (DeepLabV3+~\cite{chen2018encoder} and MaskFormer~\cite{cheng2021per}).} We leverage public checkpoints when available. In cases where a checkpoint is not available, we retrain the model using the original training hyperparameters (\eg optimizer, learning rate, momentum, and weight decay) along with the standard training schedule, which varies depending on the dataset (40k iterations for P. VOC, 80k for P. Context, and 160k for COCO). We show the performance of DeepLabV3+ in qualitative comparisons (\emph{Fully Sup.}).

\section{Additional results}

\subsection{Additional datasets}

 We evaluate our method on two new challenging datasets that contain significantly more classes, LVIS (1103 classes) and ImageNet-S (919 classes). The results are shown in Table~\ref{tab:lvis}.
We observe that our model outperforms several existing open-vocabulary baseline methods and approaches supervised models, indicating its robustness in challenging scenarios. 
\begin{table}[t]
\centering
\small
\tablestyle{5pt}{0.95}
\begin{tabular}{c|cc|cc}
Method  & OV & Sup. & \makecell{LVIS \\ (1103 classes)} & \makecell{ImageNet-S \\ (919 classes)} \\
\midrule
CLIP~\citep{radford2021learning}  & \cmark & text & 1.3 & 8.0 \\
MaskCLIP~\citep{zhou2022extract}  & \cmark & text & 4.3	& 9.1 \\
GroupViT~\citep{xu2022groupvit} & \cmark & text & 7.2 & 32.2\\
\textbf{\mname{} (Ours)}  &  \cmark & text & \textbf{8.5} & \textbf{34.9} \\
\midrule
ViT-FCN~\footnotemark  &  \xmark & GT & 9.6 &  40.4

\end{tabular}
\caption{\small \textbf{Open-vocabulary semantic segmentation results on  LVIS and ImageNet-S.} Our method demonstrates competitive performance on these challenging datasets with a significantly larger number of classes.}
\label{tab:lvis}
\end{table}

\footnotetext{\vspace{-5pt}We also tried DeepLabV3+ but failed to obtain satisfactory results.}

\begin{table*}[h]
\vspace{-0.20in}

\centering
\small
\tablestyle{4pt}{1.05}
\begin{tabular}{cc|cc|cc|c|c|c}
Method  &  Algorithm &  VL Pretrain & Pretrain data & Anno.  masks & I-T pairs & Custom model & Loss & mIoU (VOC) \\
\midrule
OpenSeg~\cite{ghiasi2022scaling} & \tablestyle{4pt}{1.0}\begin{tabular}[c]{@{}c@{}}Adapt\&Refine image-level \\ VL alignment models \end{tabular}  & \textcolor{red}{Yes (ALIGN)} &  1800M &  \textcolor{red}{Yes (COCO)} & - &  \textcolor{blue}{Not required} & image+pixel &  77.2\\

\midrule

ZegFormer~\cite{ding2022decoupling} & \tablestyle{4pt}{1.0}\begin{tabular}[c]{@{}c@{}}Directly training \\ pixel\&language alignment \end{tabular}  & \textcolor{red}{Yes (CLIP)}  &  400M & \textcolor{red}{Yes (COCO)} & - & \textcolor{blue}{Not required}  & image+pixel & 80.7 \\
\midrule

MaskCLIP~\cite{zhou2022extract} & \tablestyle{4pt}{1.0}\begin{tabular}[c]{@{}c@{}}Adapt\&Refine image-level \\ VL alignment models \end{tabular}  & \textcolor{red}{Yes (CLIP)} &  400M & \textcolor{blue}{Not required} & - & \textcolor{blue}{Not required} & image & 49.5 \\

\midrule
GroupViT~\cite{xu2022groupvit} & \tablestyle{4pt}{1.0}\begin{tabular}[c]{@{}c@{}}Extract segments \\ from language alignment \end{tabular}  & \textcolor{blue}{Not required} &  -  &  \textcolor{blue}{Not required} & 30M & \textcolor{red}{Yes (GroupViT)} & image & 77.2  \\

\midrule
\textbf{\mname{} (Ours)} & \tablestyle{4pt}{1.0}\begin{tabular}[c]{@{}c@{}}Directly training \\ pixel\&language alignment \end{tabular} & \textcolor{blue}{Not required} 
& -  & \textcolor{blue}{Not required}  & 26M & \textcolor{blue}{Not required} & image+pixel & 81.8
\end{tabular}

\caption{\small \textbf{Comparing \mname{} (Ours) with closely-related methods (OpenSeg~\citep{ghiasi2022scaling}, ZegFromer~\cite{ding2022decoupling}, MaskCLIP~\cite{zhou2022extract}, and GroupViT~\cite{xu2022groupvit}).} We conduct a comparative analysis of our method against a range of closely-related approaches, which are further detailed in Section~\ref{sec:method_compare}.}
\label{tab:diff1}
\vspace{-0.10in}
\end{table*}

\begin{figure*}[t]
    \centering
    \includegraphics[width=\linewidth]{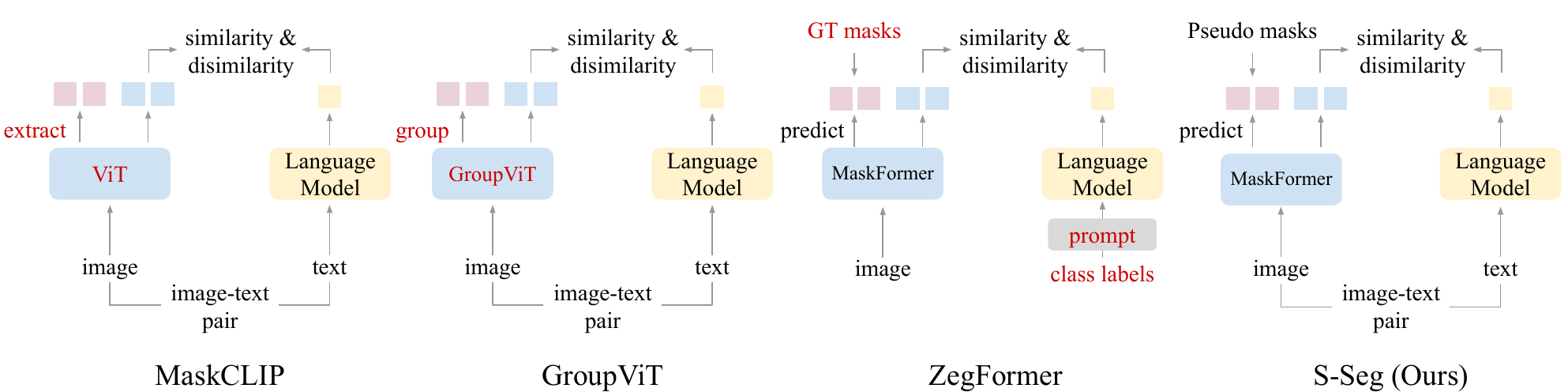}
    
    \caption{\small \textbf{Comparing \mname{} (Ours) with closely-related methods.} The components in red are those different from \mname{}.}
    \vspace{-1em}
    \label{fig:method_compare}
\end{figure*}

\subsection{Ablation results}
In Table~\ref{tab:ablations_supp}, we show numerical results corresponding to Figure~10 and 12 in the main paper. As seen from the table, scaling data and self-training provide consistent gain in performance for our model. 

\subsection{Per-category result}
Table~\ref{tab:voc_per_class} presents the mIoU results of our models and baseline methods on the Pascal VOC dataset, where each class is evaluated separately. Our models outperform GroupViT in most classes, and \mname{}+ achieves superior performance across \emph{all} categories. Our models are particularly effective at segmenting large objects such as aeroplanes, buses, and trains, with an average improvement of 11.1 compared to 2.5 for all classes. This improvement could suggest that our models benefit from the pseudo-mask generator, which works better for larger objects (which shows a 83.3\% oracle performance compared to 77.2\% for other classes). On the other hand, our self-training model performs better on categories that share consistent texture, such as cats, cows, dogs, and sheep, with an average improvement of 14.3 compared to 8.8 for all classes. This indicates that self-training can identify common features and reduce noise in the self-training labels.

\subsection{Additional visualizations}
\begin{figure*}[t]
    \centering
    \vspace{-1em}

    \includegraphics[width=.85\linewidth]{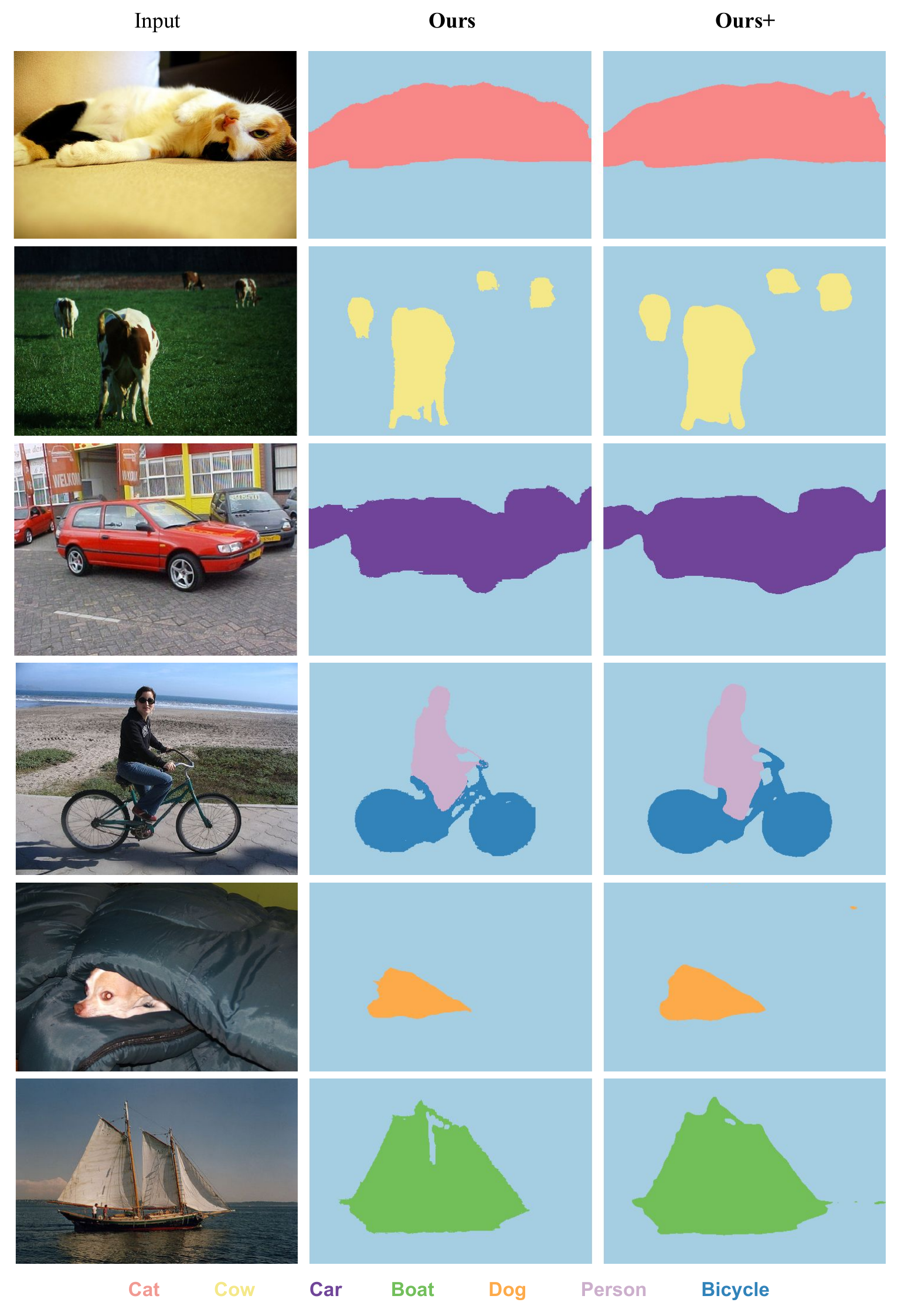}
    \caption{\textbf{Additional qualitative results of \mname{} in higher resolution (object-centric images).} Our method demonstrates robustness in dealing with challenging scenarios, such as objects with unconventional shapes and poses (row 1), images with unusual color and tone (row 2), objects of the same class but with differing colors (row 3), objects with the similar color but of different classes (row 4), concealed objects (row 5), and various other difficult situations. }
    \vspace{-1em}
    \label{fig:supp_qualitative1}
\end{figure*}

\begin{figure*}[t]
    \centering
    \includegraphics[width=.9\linewidth]{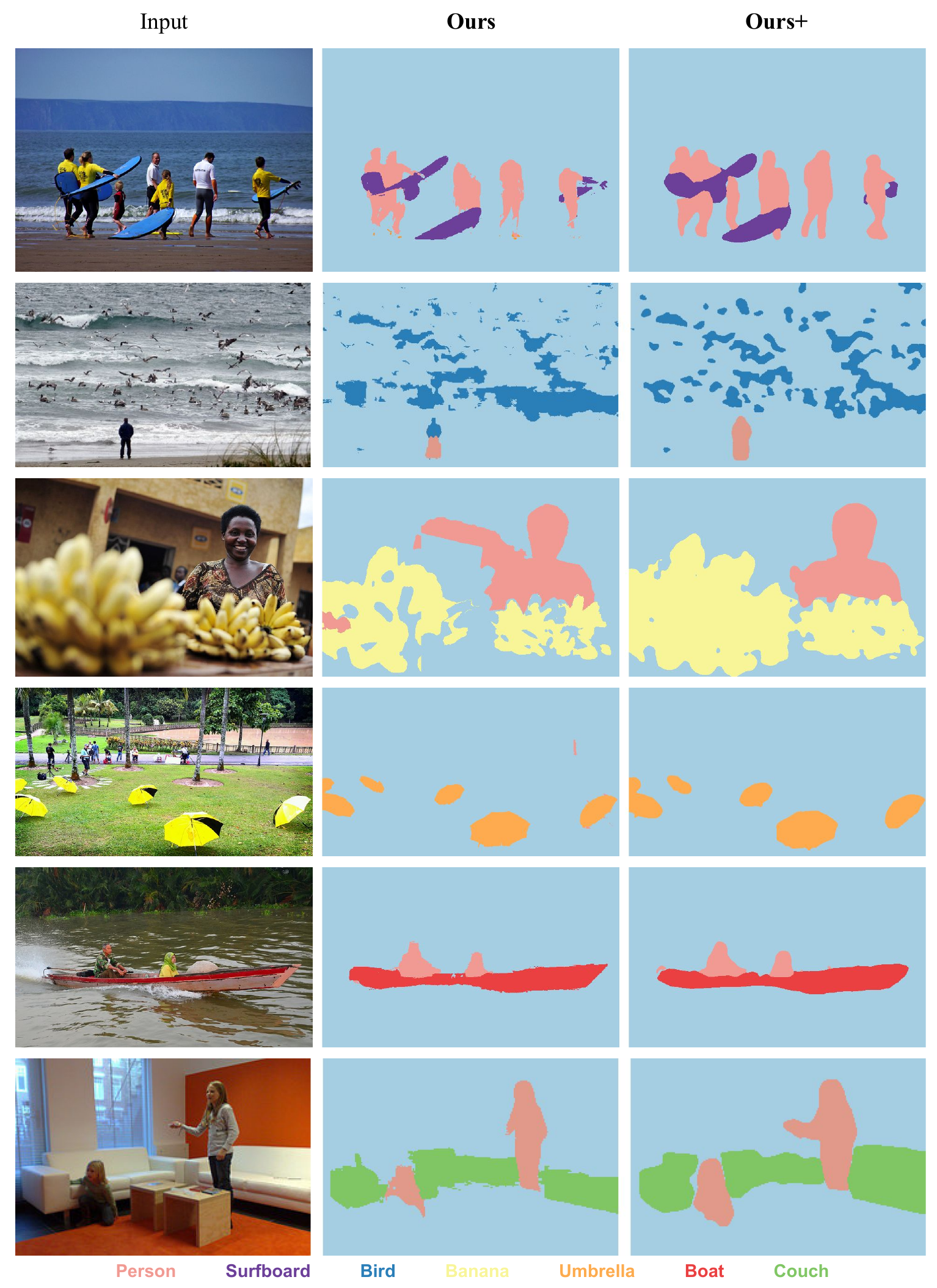}
    \caption{\textbf{Additional qualitative results of \mname{} in higher resolution (context-rich images).} Although context-rich images pose challenges in segmentation due to the presence of an increased number of small and cluttered objects, our method can still accurately segment the objects with precision.}
    \vspace{-1em}
    \label{fig:supp_qualitative2}
\end{figure*}

\begin{figure*}[t]
    \centering
    \includegraphics[width=0.99\linewidth]{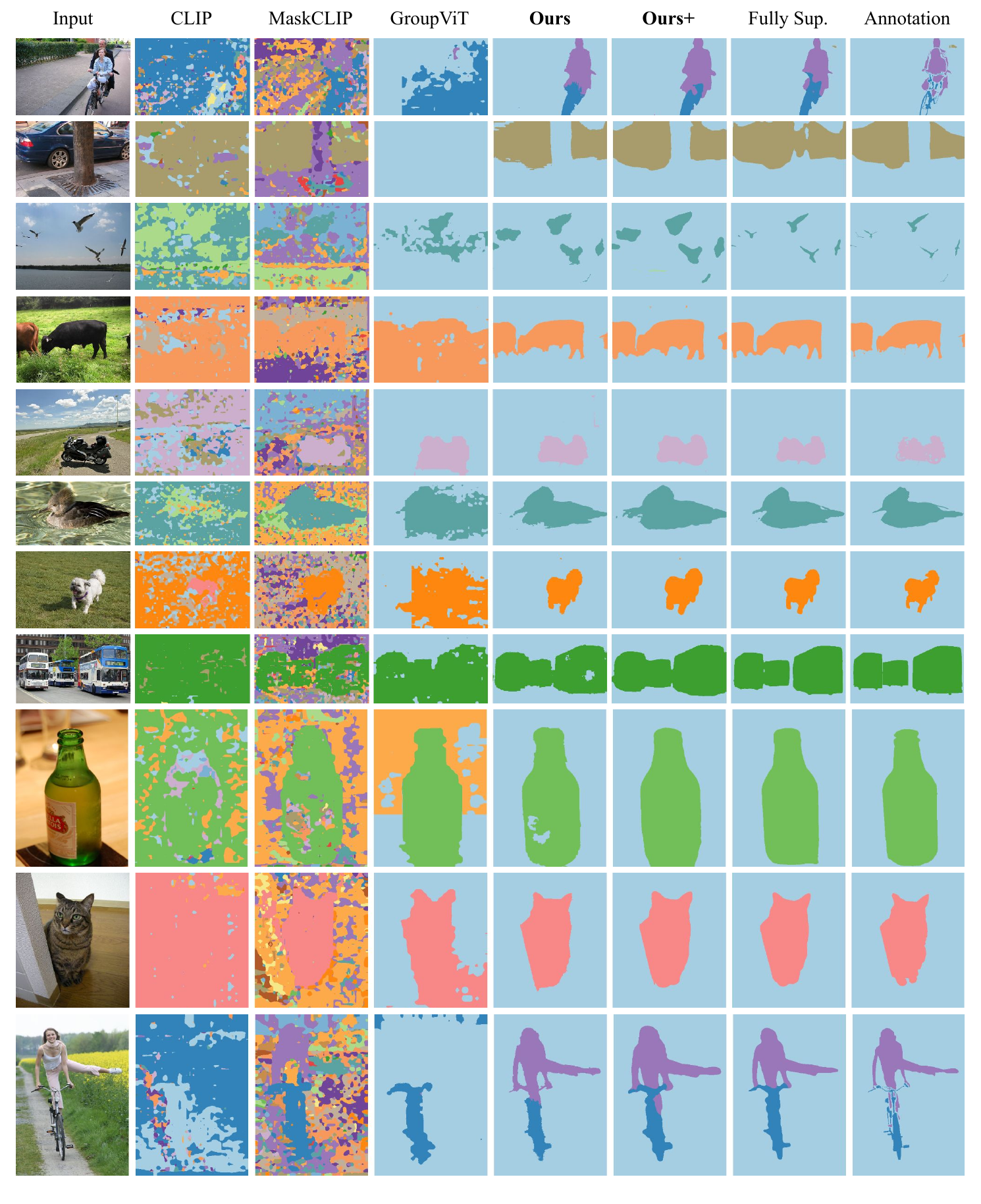}
    \caption{\textbf{Additional qualitative comparison with existing methods.}  CLIP~\cite{radford2021learning} is primarily designed for classification and does not perform well in segmentation. MaskCLIP~\cite{zhou2022extract} adapts CLIP for segmentation, although it produces noisy predictions and cannot handle background classes. GroupViT~\cite{xu2022groupvit} is a strong competitor, but it could struggle in challenging scenarios. }
    \vspace{-1em}
    \label{fig:supp_compare1}
\end{figure*}

\begin{figure*}[t]
    \centering
    \includegraphics[width=0.85\linewidth]{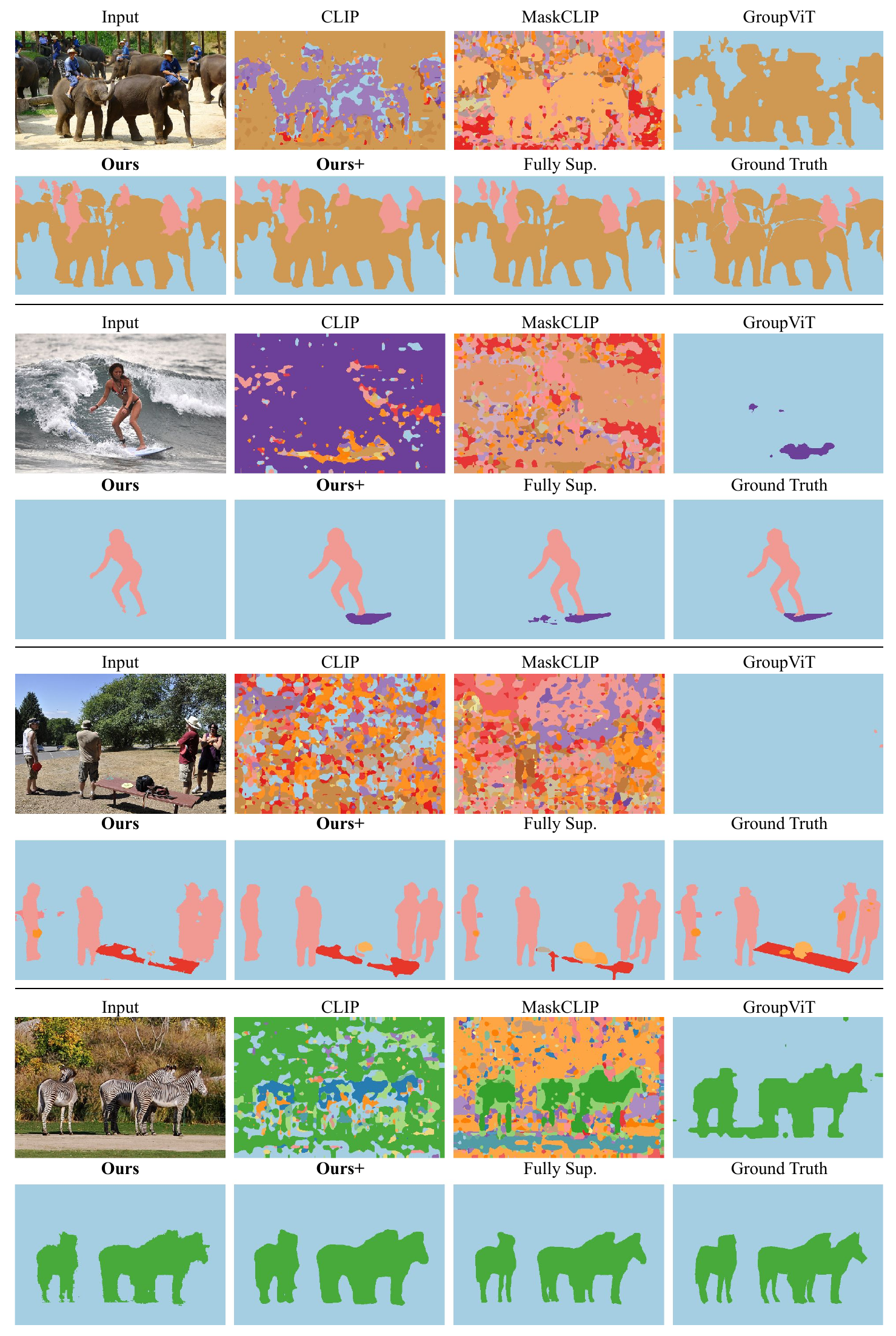}
    \caption{\textbf{Additional qualitative comparison with existing methods (continued).} }
    \vspace{-1em}
    \label{fig:supp_compare2}
\end{figure*}

Figures \ref{fig:supp_qualitative1} and \ref{fig:supp_qualitative2} present more detailed open-vocabulary segmentation results in higher resolution. As shown in the results, our approach can effectively segment object-centric images  from~\cite{everingham2009pascal} (fig.~\ref{fig:supp_qualitative1}) as well as context-rich images  from~\cite{lin2014microsoft} (fig.~\ref{fig:supp_qualitative2}) accurately. Our method can segment objects based solely on their category name, without requiring any annotations from specific target datasets during training. 
Figure~\ref{fig:supp_compare1} and~\ref{fig:supp_compare2} provide additional comparison with previous methods.

\section{Methodology Comparisons}
\label{sec:method_compare}
We present a comparative analysis of our method against several closely-related exemplary approaches. Our method serves as a connection among these methodologies. The primary similarities and differences are outlined in Table~\ref{tab:diff1}, with further discussion below.

\textbf{Relation to OpenSeg.}
OpenSeg (and similar methods, \eg LSeg~\cite{li2022language}) refines image-level models like CLIP/ALIGN by training on annotated semantic masks. The pretrained image-level model provides language alignment and utilize ground truth mask for refining pixel-level feature. In contrast, \mname{} trains directly on pixel features from pseudo-masks and learns language alignment through text. Conceptually, \mname{} offers an end-to-end alternative to OpenSeg, with the added advantage of training exclusively on image-text pairs. Our approach removes the need for the resource-intensive VL pretraining step, streamlines the learning process, and reduces the reliance on extensive supervised data.

\textbf{Relation to ZegFormer.} 
Our method can be conceptualized as a variant of "ZegFormer trained from scratch with pseudo-masks and language," albeit with notable implementation distinctions. Training with \emph{seen} ground truth masks benefits in-domain classes, but may not extend to \emph{unseen} classes. Interestingly, while our method underperforms compared to ZegFormer on \emph{seen} classes, it surpasses ZegFormer in handling \emph{unseen} classes and demonstrates superior average performance across the dataset. This suggests that our solution offers better generalization than ZegFormer, despite not utilizing CLIP, annotated masks, or pixel-wise labels. The architectural and training similarities between the two methods suggest that their integration could lead to enhanced performance, a hypothesis we leave for future exploration.

\textbf{Relation to CLIP/MaskCLIP.} 
Our method closely parallels CLIP in the image-text contrastive training paradigm and can be seen as a "CLIP with MaskFormer as the image encoder," supplemented by an additional mask supervision branch. Despite these similarities, CLIP primarily aims to learn image-level alignment, whereas \mname{} is focused on pixel-level alignment. This is evident from the fact that even with the MaskCLIP adaptation, the segmentation performance significantly lags behind that of other compared methods. This highlights the importance of incorporating both the MaskFormer and mask supervision in \mname{}.

\textbf{Relation to GroupViT.}
GroupViT and \mname{} share a similar problem setup, where both methods avoid CLIP pretraining and manual annotations. Methodologically, \mname{} resembles "GroupViT with MaskFormer as the grouping model." A key difference, however, is that GroupViT \emph{extracts} segments from a trained model, while \mname{} \emph{directly predicts} segmentation, supervised by pseudo-masks. This more explicit form of supervision allows \mname{} to leverage standard segmentation models like MaskFormer more effectively and offers a potentially simpler pathway for updates with future advancements in segmentation models.

\end{document}